\documentclass[a4paper,fleqn]{cas-dc}

\usepackage[numbers]{natbib}
\usepackage{graphicx}
\usepackage{rotating}
\usepackage{pdflscape}
\usepackage{longtable}
\usepackage{caption}
\usepackage{subcaption}
\usepackage{comment}
\usepackage{arydshln}
\usepackage{amsmath}
\usepackage{amssymb}
\usepackage{nicematrix}
\usepackage{makecell}
\usepackage{todonotes}
\usepackage{xtab}
\usepackage{booktabs}
\usepackage{float}
\usepackage{threeparttable}

\usepackage{pifont}
\newcommand{\cmark}{\ding{51}}%
\newcommand{\xmark}{\ding{55}}%

\def\tsc#1{\csdef{#1}{\textsc{\lowercase{#1}}\xspace}}
\tsc{WGM}
\tsc{QE}
\tsc{EP}
\tsc{PMS}
\tsc{BEC}
\tsc{DE}

\newcommand{\TP}{\text{TP}}
\newcommand{\FP}{\text{FP}}
\newcommand{\TN}{\text{TN}}
\newcommand{\FN}{\text{FN}}
\newcommand{\precision}{\text{precision}}
\newcommand{\recall}{\text{recall}}
\newcommand{\spfty}{\text{specificity}}
\newcommand{\accy}{\text{accuracy}}
\begin{document}
\let\WriteBookmarks\relax
\def\floatpagepagefraction{1}
\def\textpagefraction{.001}
\shorttitle{SLR of automated ICD coding and classification systems using discharge summaries}
\shortauthors{Rajvir Kaur et~al.}

%%% Actual title
%\title [mode = title]{This is a specimen $a_b$ title}                      
%\tnotemark[1,2]

%\tnotetext[1]{This document is the results of the research
%project funded by the National Science Foundation.}

%\tnotetext[2]{The second title footnote which is a longer text matter
%to fill through the whole text width and overflow into
%another line in the footnotes area of the first page.}
%%%
\title [mode = title]{A Systematic Literature Review of Automated ICD Coding and Classification Systems using Discharge Summaries}  

\author[1]{Rajvir Kaur} [orcid=0000-0002-6071-1264]                      
\cormark[1]
\ead{18531738@student.westernsydney.edu.au}
%\ead[url]{www.cvr.cc, cvr@sayahna.org}

\credit{Conceptualization of this study, Methodology, Software}

\address[1]{School of Computer, Data and Mathematical Sciences, Western Sydney University, Australia}

\author[1]{Jeewani Anupama Ginige} [orcid=0000-0002-6695-6983]
\ead{j.Ginige@westernsydney.edu.au}
\credit{Provide medical domain knowledge, Validation, Writing-review and editing, Supervision}

\author[1]{Oliver Obst} [orcid=0000-0002-8284-2062]
\ead{O.Obst@westernsydney.edu.au}
%\ead[URL]{www.sayahna.org}
\credit{Provide knowledge in Artificial Intelligence domain, Writing-review and editing, Validation, Supervision}

%\address[2]{School of Computer, Data and Mathematical Sciences, Western Sydney University, Australia}

\cortext[cor1]{Corresponding author}
%\cortext[cor2]{Principal corresponding author}
%\fntext[fn1]{This is the first author footnote. but is common to third
  %author as well.}
%\fntext[fn2]{Another author footnote, this is a very long footnote and
  %it should be a really long footnote. But this footnote is not yet
  %sufficiently long enough to make two lines of footnote text.}

%\nonumnote{This note has no numbers. In this work we demonstrate $a_b$
 %the formation Y\_1 of a new type of polariton on the interface
  %between a cuprous oxide slab and a polystyrene micro-sphere placed
  %on the slab.
  %}

\begin{abstract}
\noindent \textbf{Background:} Codification of free-text clinical narratives have long been recognised to be beneficial for secondary uses such as funding, insurance claim processing and research. The current scenario of assigning codes is a manual process which is very expensive, time-consuming and error prone. In recent years, many researchers have studied the use of Natural Language Processing (NLP), related machine learning and deep learning methods and techniques to resolve the problem of manual coding of clinical narratives and to assist human coders to assign clinical codes more accurately and efficiently.

\noindent \textbf{Objective:} The main objective of this systematic literature review is to provide a comprehensive overview of automated clinical coding system that utilises appropriate NLP, machine learning and deep learning methods and techniques to assign ICD codes to discharge summaries. 

\noindent \textbf{Method:} We have followed the Preferred Reporting Items for Systematic Reviews and Meta-Analyses (PRISMA) guidelines and conducted a comprehensive search of publications from January, 2010 to December 2020 in four high quality academic databases- PubMed, ScienceDirect, Association for Computing Machinery (ACM) Digital Library, and the Association for Computational Linguistics (ACL) Anthology. 

\noindent \textbf{Result:} We reviewed 7,556 publications; 38 met the inclusion criteria. This review identified: 6 datasets having discharge summaries (2 publicly available, 4 acquired from hospitals); 14 NLP techniques along with some other data extraction processes, different feature extraction and embedding techniques. The review also shows that there is a significant increase in the use of deep learning models compared to machine learning. To measure the performance of classification methods, different evaluation metrics are used. Lastly, future research directions are provided to scholars who are interested in automated ICD code assignment system.

\noindent \textbf{Conclusion:} Efforts are still required to improve ICD code prediction accuracy, availability of large-scale de-identified clinical corpora with the latest version of the classification system. This can be a platform to guide and share knowledge with the less experienced coders and researchers.

\end{abstract}

%\includegraphics{figs/grabs.pdf}

\begin{highlights}
\item A systematic literature review focus on automated ICD code assignment using discharge summaries was conducted.
\item A total of 38 studies published between January 2010 and December 2020 were selected and analysed.
\item We review computerised systems that employ Artificial Intelligence, Machine Learning, Deep Learning and Natural Language Processing for assigning ICD codes to discharge summaries.
\item We highlighted limitations in existing studies and discussed open research challenges.
\end{highlights}


\maketitle

\section{Introduction}

Documentation related to an episode of care of a patient, commonly referred to as a medical record, contains clinical findings, diagnoses, interventions, laboratory test details and medication details which are invaluable information for clinical decisions making. To carry out meaningful statistical analysis, these medical records are converted into a special set of codes which are called \textit{clinical codes} as per clinical coding standards set by the World Health Organisation (WHO). The International Classification of Diseases (ICD) is a special set of alphanumeric codes, assigned to an episode of care of a patient, based on which reimbursement is done in some countries~\cite{kaur2018}. Clinical codes are assigned by trained professionals, known as \textit{clinical coders}, who have a sound knowledge of medical terminologies, clinical classification systems, and coding rules and guidelines. The current scenario of assigning clinical codes is a manual process which is very expensive, time-consuming, and error-prone \cite{Xie2018}. The wrong assignment of codes leads to issues such as reviewing of the whole process, financial losses, increased labour costs as well as delays in reimbursement process. The coded data is not only used by insurance companies for the reimbursement purposes, but also by government agencies and policy makers to analyse healthcare systems, justify investments done in the healthcare industry and plan future investments based on these statistics ~\cite{kaur2018}.

With the transition from ICD-9 to ICD-10 in 1992, the number of codes increased from 3,882 to approximately 70,000, which further makes manual coding a non-trivial task~\cite{subotin2014}. Moreover, the manual assignment of codes is a complex process due to the continuous evolution of rules. As an example, clinical coding competency guidelines in Australia\footnote{https://www2.health.vic.gov.au/health-workforce/health-information-workforce/clinical-coding-workforce} require that an entry-level clinical coder is able to code a minimum of 5 clinical records per hour with 85\% accuracy. However, in practice, on average, a clinical coder codes 3 to 4 clinical records, resulting in 15-42 records per day depending on the experience and efficiency of the human coder~\cite{kaur2018, santos2008}. The cost incurred in assigning clinical codes and the follow up corrections are estimated to be 25 billion dollars per year in the United States~\cite{Farkas2008,Xie2018}. There are several reasons behind the wrong assignment of codes. First, the assignment of ICD codes to a patient's records could be erroneous due to the subjective nature of human perception~\cite{Rajvir2019-distributed}. Second, the manual process of assigning codes is a tedious task which leads to inability to locate critical and subtle findings due to fatigue. Third, in many cases, physicians or doctors often use abbreviations or synonyms, which causes ambiguity~\cite{Xie2018}.

\citet{butler2017} highlights major future challenges that health information management practitioners and academics will face with an ageing workforce, where more than 50\% of the workforce is aged 45 years or older.  \citet{Huang2019} show that more than 80\% of health record data is in an unstructured form which acts as a barrier in an automated clinical decision making process. Unstructured clinical text includes clinical notes, surgical records, discharge summaries, radiology reports, and pathology reports. This unstructured text contains a lot of valuable information but lacks common structural frameworks and may contain errors, such as spelling errors, grammatical errors, and semantic ambiguities, which further increases the complexity of data processing and analysis~\cite{Wencheng2018}. There are various other factors that hinder the clinical decision making process:
\begin{enumerate}
    \item \textit{Idiosyncrasies of medical language}: Free text clinical notes are rife with obscure vocabulary, non-standard syntax, and ambiguous abbreviations. These free text clinical notes are often typed hurriedly, and thus, contain many spelling and grammatical errors~\cite{Catling2018}. Moreover, synonyms for clinical concepts are also used interchangeably and negating expressions are placed distantly from the negated concept \cite{Chapman2001}. 
    
    \item \textit{Scarcity of electronic health records (EHRs)}: This is a long-standing barrier to automated coding as many hospitals are still using paper records, which limits the availability of training data. There is limited adoption of structured EHRs in developing countries, which leaves clinicians with no choice but to resort to manual consumption of available clinical notes for decision making. Sometimes crucial information about a patient is mostly lost when transcribed into structured EHRs~\cite{GANGAVARAPU2020}.  
    
   \item \textit{Label-space problem}: Many disease ontologies contain tens of thousands of labels, and their distribution is highly imbalanced in most datasets, with many absent labels for rare diseases. Many research studies have used k-most frequently occurring diseases (or labels) for training and discarded the least occurring labels in order to achieve higher accuracy of their algorithm~\cite{Catling2018}. However, this type of negligence would be unacceptable in real healthcare environments where many rare diseases have serious sequelae when neglected. 
   
   \item \textit{Requirement of large amount of training data}:  It is quite rare to find freely available medical repositories that contains clinical records of a patient. To achieve high performance of a model, it is necessary to have a large amount of training data as machine learning models learn from experiences \cite{kaur2018}.
 
\end{enumerate}

To reduce coding errors and cost, there is a need for an automated clinical coding system, commonly referred to as computer-assisted coding system that will overcome the manual coding challenges and assist human coders to assign correct clinical codes more quickly and accurately. Several previous studies have addressed the automated ICD coding systems, but currently not widely used, most likely because the systems are still in development and their performance in a real time scenario is unproven~\cite{Stanfill2010}. The Automated coding methods make use of Artificial Intelligence (AI) techniques to convert unstructured clinical text to structured text without human interaction~\cite{RajvirThesis2018}. In addition, a wide range of applications in the biomedical domain uses natural language processing (NLP) techniques to manage large volumes of text data by extracting relevant information in a timely manner. 

\section*{Our Contribution}

 There are only three systematic literature reviews available for clinical coding and classification systems \cite{Burns, campbell2001systematic, Stanfill2010}. Two studies \cite{Burns, campbell2001systematic} conducted a systematic review to measure the accuracy of routinely collected  hospital discharge data in the United Kingdom. The primary objective of these two studies were to identify and investigate the accuracy of hospital episode data and to investigate factors affecting variation in coding. The studies did not focus on automated ICD code assignment using any computer application. \citet{Stanfill2010} conducted a systematic literature review of studies that use computer application to automatically generate clinical codes or classification from free-text clinical documents. This review includes the studies published prior to March 2009. Apart from that, \citet{Mujtaba2019} focused on clinical text classification studies published from January 2013 to January 2018. However, this systematic literature review investigates all types of clinical reports including pathology reports, radiology reports, autopsy reports, death certificates and other medical reports for text classification purpose, rather than automated ICD coding and classification purpose. To the best of our knowledge, no systematic literature review has recapitulated the existing studies on automated ICD code assignment on discharge summaries in the last one decade. Our contribution is to conduct a systematic literature review on automated ICD code assignment using discharge summaries published from January 2010 to December 2020. Within the broader scope of this review, the work aims to address the following research questions: 
 
 \begin{enumerate}
     \item What are the close ended quality assessment questions that make the study suitable for systematic literature review?
     \item What are the different datasets available for ICD code assignment of discharge summaries?
     \item What are the different computerised algorithms available to assign automated ICD codes?
     \item What are the different evaluation metrics used in the studies to evaluate the performance of automated ICD code assignment systems?
     \item Which model gives best performance in assigning ICD codes to discharge summaries?
     \item What are the future research directions in automated ICD coding task?
 \end{enumerate}
 
\begin{table*}[t]
\caption{Comparison between ICD-9-CM, ICD-10-CM/ICD-10-PCS and ICD-10-AM/ACHI}
\label{tab:compareICD}
\centering
\begin{tabular}{p{5cm}p{5cm}p{5cm}}
\toprule
\textbf{ICD-9-CM} & \textbf{ICD-10-CM/ICD-10-PCS}& \textbf{ICD-10-AM/ACHI} \\ 
\midrule

 The first character is alpha (E and V only) or a numeric & The first character is Alphanumeric & The first character is Alphanumeric \\ \addlinespace[1mm]
 
Approx. 13,500 diagnoses & Approx.	70,000 diagnoses &	Approx. 16,953 disease codes \\
 & & 2,825 morphology codes \\ \addlinespace[1mm]
 
 Valid codes have 3 to 5 characters & Valid codes have 3 to 7 characters & Valid codes have 3 to 5 characters \\ \addlinespace[1mm]
 Approx. 4,000 procedures & Approx.	72,000 procedures	& Approx. 6,248 ACHI codes\\ \addlinespace[1mm]
Valid codes have 4-digit maximum  &	Valid codes have 7-digit maximum  & Valid codes have 7-digit maximum	\\ \addlinespace[1mm]
Decimal used after the second digit & No decimals used & No decimals used \\ 
\bottomrule
\end{tabular}
\end{table*}

\section{Background}
The history of medical coding started as an attempt to avoid the black death known as the bubonic plague, caused by the bacteria \textit{Yerisina pestis}, which arrived in Sicily via ship rats in the year 1347. The outbreaks of plague continued in Europe throughout the next 3 centuries. In the year 1532, the systematic collection of data on causes of death known as \textit{The London Bills of Mortality} began and these data were published weekly. Causes of death found in the Bills included diseases such as jaundice, smallpox, rickets, spotted fever, and plague. 
In the year 1665, John Graunt, a London merchant, published \textit{Reflections on the Weekly Bills of Mortality} to examine the deaths from plague in the context of all other causes of mortality. Later on in the year 1839, Dr.\ William Farr prepared a classification system which was primarily based on the anatomical site and consisted of 138 rubrics \cite{Moriyama2011}. After the death of Dr.\ Farr in 1883, Jacques Bertillon, a French statistician, prepared a revised list that was adopted by the International Statistical Institute in 1893. The Bertillon Classification was the first standard system implemented internationally to record causes of death known as \textit{the International Classification of Diseases}. Delegates from 26 countries adopted the Bertillon Classification (ICD-1) in 1900 and subsequent revisions occurred since 1920. After Bertillon's death in 1922, interest grew in using the classification to categorise not only causes of mortality, but also causes of morbidity.

In April 1948, at the Sixth Decennial Revision Conference in Paris, the WHO approved a comprehensive list for classification of causes of illness (morbidity), as well as causes of deaths (mortality). The ``Manual of the International Statistical Classification of Diseases, Injuries, and Causes of Death" is generally known as ICD.  With the eighth revision, the United States developed its own version, known as ICDA-8 (ICD-Adapted), due to disagreement over the circulatory section. With the ninth revision in 1976, the United States adopted a clinical modification of the international version (ICD-9-CM) and used it until October 2015. Although ICD-10 was endorsed by the WHO in 1990 to classify mortality data from death certificates, ICD-9 was used for all other purposes, including billing and reimbursement \cite{Moriyama2011}. Many countries extended the ICD-10 classification system to make it suitable for their country specific reporting purposes \cite{kaur2018}. For example, ICD-10-CM (Clinical Modification) is used in the USA, ICD-10-CA (Canadian Modification) is used in Canada, ICD-10-GM (German Modification) in Germany, and ICD-10-AM (Australian Modification) is used in Australia along with 15 other countries including  Ireland, Singapore, and Saudi-Arabia \cite{cumerlato2010, kaur2018}. Twenty-six years after the introduction of ICD-10, in 2018, the next generation of classification ICD-11 was put forward to WHO general assembly for approval and was released in May 2019, but it is not yet implemented in the hospital settings \cite{kaur2018, RajvirThesis2018}. ICD-11 increases the complexity by introducing a new code structure, with a new chapter on X-Extension Codes, dimensions of external causes (histopathology, consciousness, temporality, and etiology), and new chapters on sleep-awake disorder, conditions related to sexual health, and traditional medicine conditions~\cite{Jenny2014,JapanWHO2016,Reed2016}.

\subsection{Comparison of different classification systems}
Since the introduction of ICD-10 in 1992, many countries have modified the WHO's ICD-10 classification system to suit their country specific reporting purpose. There are a few major differences between the US and Australian classification systems. Firstly, there are a few additional ICD-10-AM codes that are more specific (approximately $4,915$ codes) and are coded in Australia along with 15 other countries that use Australian classification system as their national classification system. For example, in the US system, \textit{`contact with venomous spiders'} is coded as $X21$, whereas in Australia, there is more specificity by adding fourth character level in identifying the type of the spider \cite{Rajvir2019-distributed}.

There are $12\%$ of ICD-10-AM specific codes that do not exist in ICD-10-CM, ICD-9-CM or any other classification system. Secondly, countries that have developed their own national classification system use different coding practices. For example, in the US, Pulmonary oedema is coded as $J81$, whereas in Australia, to assign code for  Pulmonary oedema, there is ACS rule 0920 which says, ``When \textit{acute pulmonary oedema} is documented without further qualification about the underlying cause, assign \textit{I50.1 Left ventricular failure}''. Table~\ref{tab:compareICD} shows the comparison of ICD-9-CM, ICD-10-CM /ICD-10-PCS, and ICD-10-AM/ACHI codes.

\section{Methods}

We followed the Preferred Reporting Items for Systematic Reviews and Meta-Analyses (PRISMA) recommendations for reporting in systematic reviews \cite{PRISMA}. After reviewing the PRISMA guidelines, we structured our review into two phase:

\begin{figure*}
    \centering
    \includegraphics[width=18cm, height=18cm]{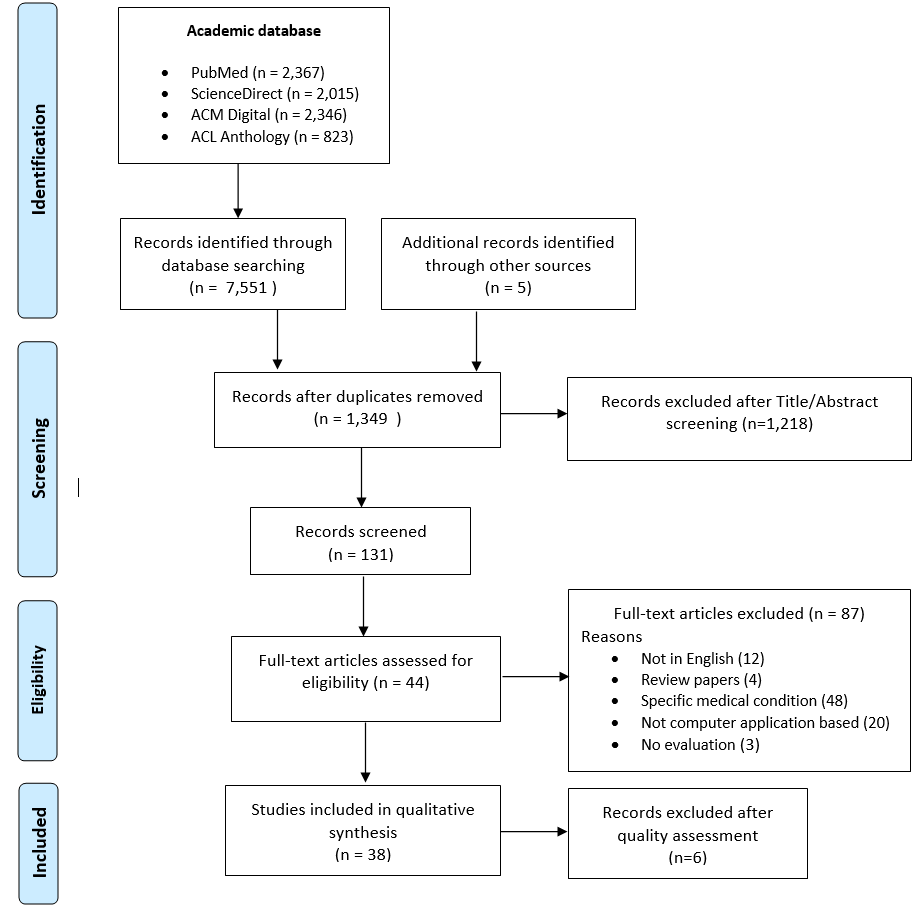}
    \caption{PRISMA Flow Diagram}
    \label{fig:flowdia}
\end{figure*}

\begin{enumerate}
    \item Search strategy phase
    \item Data extraction phase
\end{enumerate}

\subsection{Search strategy phase}
The search strategy phase includes data sources, formulation of the search keywords and search queries, screening and selection criteria, and quality assessment of the retrieved publications. This phase was designed to identify all potential relevant publications related to the automated clinical coding and classification systems that would leverage on NLP, machine learning and deep learning techniques. 

\subsubsection{Data sources and search strategies} 
We conducted a comprehensive search of several databases for publications from January 1, 2010 to December 31, 2020. Since January 2010 onwards, no comprehensive systematic literature review has recapitulated on automated clinical coding using discharge summaries. In this review, the studies were retrieved from four high quality academic databases- PubMed, ScienceDirect, Association for Computing Machinery (ACM) Digital Library, and the Association for Computational Linguistics (ACL) Anthology. We looked for publications which include conference proceedings and journal articles written in English, and excluded those in the form of editorial, letter, note or comment. Table~\ref{tab:inclusion} shows a list of study selection criteria.

\begin{table*}
\caption{List of the inclusion and exclusion criteria}
\label{tab:inclusion}
\centering
\begin{tabular}{cp{15cm}}
\toprule
\textbf{S.No} & \textbf{Inclusion Criteria} \\
\midrule
1. & Study must be published between January 1, 2010 and December 31, 2020.\\
2. & Study must have used discharge summary or other medical reports along with discharge summary as a dataset and must be written in the English language.\\
3. & Study must have used clinical reports or medical documents for automated ICD code assignment. \\
4. & Study should perform automated ICD code assignment using computer applications such as NLP, machine learning and deep learning techniques. \\
5. & Study must have evaluated the performance of the proposed system using standard evaluation metrics. 

\\
\textbf{S.No} & \textbf{Exclusion Criteria} \\ \hline
1. & Study using reports other than discharge summary such as radiology report, pathology report, death certificate, autopsy, surgical or laboratory report.\\
2. & Study has not defined any clinical coding or clinical classification system.\\
3. & Study assigning clinical codes on one specific condition (disease or procedure). \\
3. & Study focused on medical image classification \\
4. & Study has not evaluated the performance of automated coding and classification system.\\
5. & Study focused on medical text snippets. \\
6. & Study has used the classification of time series data in the medical field such as EEG signals and nothing to deal with text classification \\

\bottomrule
\end{tabular}
\end{table*}

\subsubsection{Formulation of keywords and search queries}
To perform the search query, each keyword is paired using OR operators, whereas the concepts are paired using AND operator. Table~\ref{tab:keyword} shows the keywords used to perform search query. 

\begin{table*}
\caption{Searched keywords using different concepts}
\label{tab:keyword}
\centering
\begin{tabular}{p{6cm}p{10cm}}
\toprule
\textbf{Concepts} & \textbf{Keywords} \\
\midrule
Concept 1: Keywords related to classification domain & ICD code classification OR ICD code assignment OR text classification OR clinical text classification OR automated text classification OR text categorisation OR computer assisted coding OR computer assisted clinical coding  \\ \addlinespace[1mm]
Concept 2: Keywords related to medical documents & medical report OR clinical report OR clinical narratives OR electronic health records OR free text clinical report OR discharge summary  \\ \addlinespace[1mm]

\textbf{Search Query} & \textbf{(Concept 1) AND (Concept 2)} \\ 

\bottomrule
\end{tabular}
\end{table*}

\subsubsection{Screening and selection criteria}
Based on the search query, the publications retrieved from each database were stored in EndNote X9 (Thomson Reuters) reference management software and the \textit{Find Duplicates} function was used to review and delete duplicates. Some manual deletion was also performed. After removing the duplicates, the remaining 1,349 publications were screened based on the titles and abstracts to determine if the study is relevant to our review. After the first screening, 131 publications were selected. In the second screening, full text PDF files were obtained using the EndNote \textit{Find Full Text} feature. The full texts that could not be found or obtained because of access restrictions were then requested and attached manually to the list. The third screening was performed to review the full text of the publications. Each paper's \textit{Methodology or Experimental Work} section was reviewed properly to determine if it meets the review criteria. During this screening phase, 87 publications were removed as they did not meet the selection criteria and the remaining 44 studies were eligible for full text assessment.

\subsubsection{Quality assessment of the retrieved publications}

The quality assessment of the selected publications was one of the essential steps to find out whether or not the study is suitable to address our review objectives. To perform the quality assessment, we formulated a checklist of ten close-ended questions as shown in Table \ref{tab:quality}, the answers to which can be either \textit{Yes} or \textit{No} and where each question carries weight of ``1". A threshold of 7 was set to include study in the review. During the quality assessment process, 6 studies were excluded as they did not obtain a score of 7. Hence, this review involves 38 selected studies. Table~\ref{tab:qualityassessment} shows the quality assessment criteria score of the selected studies.

\begin{table*}
\caption{List of quality assessment questions}
\label{tab:quality}
\centering
\begin{tabular}{cp{12cm}}
\toprule
\textbf{S.No.} & \textbf{Question} \\
\midrule
1. & Are research objectives clearly defined? \\
2. & Is research methodology well-defined?\\ 
3. & Is the train and test data source clearly defined? \\
4. & Are the data pre-processing techniques clearly defined and their selection justified? \\
5. & Are the feature extraction techniques or feature engineering clearly described and justified? \\ 
6. & Are the classifiers clearly described? \\
7. & Does the study perform the comparison with the existing baseline models? \\
8. & Is the performance of the system evaluated and results properly interpreted and discussed? \\
9. & Does the study performs an ablation study? \\
10. & Does the conclusion reflect the research findings? \\

\bottomrule
\end{tabular}
\end{table*}

\begin{table*}
\caption{Summary of quality assessment studies}
\label{tab:qualityassessment}
\centering

\begin{tabular}{p{0.5cm}p{4cm}p{0.7cm}p{0.7cm}p{0.7cm}p{0.7cm}p{0.7cm}p{0.7cm}p{0.7cm}p{0.7cm}p{0.7cm}p{0.7cm}p{0.9cm}}
\hline
\toprule
\textbf{Year} & \textbf{Study} & \textbf{Q1} &\textbf{Q2}  & \textbf{Q3} &\textbf{Q4}  & \textbf{Q5} & \textbf{Q6} & \textbf{Q7} & \textbf{Q8} & \textbf{Q9} & \textbf{Q10} & \textbf{Score} \\ \hline

\midrule
2013 & \citet{Perotte2013} & \cmark & \cmark & \cmark & \cmark &\cmark & \cmark & \xmark & \cmark & \xmark &\cmark & 8\\

2014 & \citet{Marafino2014} & \cmark & \cmark & \cmark & \cmark  & \cmark & \cmark & \xmark & \cmark & \xmark & \cmark & 8\\ 

2014 & \citet{subotin2014} & \cmark &\cmark  & \xmark &\cmark  & \cmark & \cmark & \xmark & \cmark & \xmark & \cmark & 7 \\

2015 & \citet{Kavuluru2015} & \cmark & \cmark & \cmark & \cmark & \cmark & \cmark & \xmark & \cmark & \xmark & \cmark & 8 \\

2016 & \citet{Ayyar2016} & \cmark & \cmark  & \cmark & \cmark  & \cmark & \cmark & \xmark & \cmark & \xmark & \cmark & 8\\

2017 & \citet{Prakash2017} & \cmark & \cmark & \cmark & \cmark & \cmark & \cmark & \xmark & \cmark & \xmark & \cmark & 8 \\

2017 & \citet{Chin2017} & \cmark & \cmark & \cmark & \cmark & \cmark & \cmark & \xmark & \cmark & \xmark & \cmark & 8 \\

2017 & \citet{Berndorfer2017} & \cmark & \cmark & \cmark & \cmark & \cmark & \cmark & \xmark & \cmark & \xmark & \cmark & 8\\

2018 & \citet{Amoia2018} & \cmark & \cmark & \cmark & \cmark & \cmark & \cmark & \xmark & \cmark & \xmark & \cmark & 8  \\

2018 & \citet{Catling2018} & \cmark & \cmark & \cmark & \cmark & \cmark & \cmark & \cmark & \cmark & \xmark & \cmark & 9  \\

2018& \citet{Baumel2018} & \cmark & \cmark & \cmark & \cmark & \cmark & \cmark & \cmark & \cmark & \xmark & \cmark & 9   \\

2018 & \citet{Mullenbach2018} & \cmark & \cmark & \cmark & \cmark & \cmark & \cmark & \cmark & \cmark & \xmark & \cmark & 9\\

2018 & \citet{Samonte2018}  & \cmark & \cmark & \cmark & \cmark & \cmark & \cmark & \cmark & \cmark & \xmark & \cmark & 9\\

2018 & \citet{Xie2018} & \cmark & \cmark & \cmark & \cmark & \cmark & \cmark & \cmark & \cmark & \cmark & \cmark & 10 \\

2018 & \citet{rios-kavuluru-2018-shot}& \cmark & \cmark & \cmark & \cmark & \cmark & \cmark & \cmark & \cmark & \xmark & \cmark & 9 \\

2018 & \citet{kaur2018} & \cmark & \cmark & \cmark & \cmark & \cmark & \cmark & \xmark & \cmark & \xmark & \cmark & 8 \\

2019 & \citet{Zeng2019} & \cmark & \cmark & \cmark & \xmark & \cmark & \cmark & \cmark & \cmark & \xmark & \cmark & 8 \\

2019 & \citet{Rajvir2019} & \cmark & \cmark & \cmark & \cmark & \cmark & \cmark & \xmark & \cmark & \xmark & \cmark & 8\\

2019 & \citet{Xie2019} & \cmark & \cmark & \cmark & \cmark & \cmark & \cmark & \cmark & \cmark & \cmark & \cmark & 10 \\ 

2019 & \citet{Falis2019}  & \cmark & \cmark & \cmark & \cmark & \cmark & \cmark & \cmark & \cmark & \cmark & \cmark & 10  \\

2019 & \citet{Huang2019}  & \cmark & \cmark & \cmark & \cmark & \cmark & \cmark & \cmark & \cmark & \xmark & \cmark & 9 \\

2019 & \citet{Min2019} & \cmark & \cmark & \cmark & \xmark & \cmark & \cmark & \cmark & \cmark & \xmark & \cmark & 8\\

2019 & \citet{Rios2019} & \cmark & \cmark & \cmark & \cmark & \cmark & \cmark & \cmark & \cmark & \xmark & \cmark & 9 \\

2019 & \citet{Henning2019}  & \cmark & \cmark & \cmark & \cmark & \cmark & \cmark & \cmark & \cmark & \xmark & \cmark & 9 \\

2019 & \citet{Xu-Keyang2019} & \cmark &\cmark  & \cmark &\cmark  & \cmark & \cmark & \xmark & \cmark & \xmark & \cmark & 8 \\

2019 & \citet{Du2019} & \cmark & \cmark  & \cmark &\cmark  & \cmark & \cmark & \cmark & \cmark & \xmark &\cmark & 9 \\

%2019 & \citet{Wiegreffe2019} & 

2020 & \citet{Cao2020}  & \cmark & \cmark & \cmark & \xmark & \cmark & \cmark & \cmark & \cmark & \cmark & \cmark & 9   \\

2020& \citet{Donglin2020}  & \cmark & \cmark & \cmark & \cmark & \cmark & \cmark & \cmark & \cmark & \xmark & \cmark & 9 \\

2020 & \citet{Thanh2020} & \cmark & \cmark & \cmark & \cmark & \cmark & \cmark & \cmark & \cmark & \cmark & \cmark & 10 \\

2020 & \citet{Song2020} &  \cmark & \cmark & \cmark & \cmark & \cmark & \cmark & \cmark & \cmark & \cmark & \cmark & 10 \\

2020 & \citet{Aaron2020} & \cmark & \cmark & \cmark & \cmark & \cmark & \cmark & \cmark & \cmark & \xmark & \cmark & 9 \\

2020 & \citet{Mascio2020-comparative} & \cmark & \cmark & \cmark & \cmark & \cmark & \cmark & \xmark & \cmark & \xmark & \cmark & 8 \\

2020 & \citet{LiFei2020} & \cmark & \cmark & \cmark & \cmark & \cmark & \cmark & \cmark & \cmark & \xmark & \cmark & 9  \\

2020 & \citet{Teng2020} & \cmark &\cmark  & \cmark &\cmark  & \cmark & \cmark & \cmark & \cmark & \cmark & \cmark & 10\\

2020 & \citet{Moons2020} & \cmark &\cmark  & \cmark & \cmark  & \cmark & \cmark & \cmark & \cmark & \xmark & \cmark & 9\\

2020 & \citet{Shaoxiong2020-dilated} & \cmark &\cmark  & \cmark &\cmark  & \cmark & \cmark & \cmark & \cmark & \xmark & \cmark & 9\\ 

2020 & \citet{Chung2020} & \cmark &\cmark  & \cmark &\cmark & \cmark & \cmark & \cmark & \cmark & \xmark & \cmark & 9\\ 

2020 & \citet{Zachariah2020-bert-xml} & \cmark &\cmark  & \cmark &\cmark  & \cmark & \cmark & \cmark & \cmark & \xmark & \cmark & 9 \\ 

%2020 & \citet{CASCADENET2020} & \cmark & \cmark & \cmark & \cmark & \cmark & \cmark & \cmark & \cmark & \xmark & \cmark & 9 \\

%2020 & \citet{Schafer2020multilingual} & \cmark & \cmark & \cmark & \cmark & \cmark & \cmark & \cmark & \xmark & \xmark & \cmark & 8 \\
\bottomrule
\end{tabular}
\end{table*}

\subsection{Data extraction phase}
The data extraction phase includes review of the selected publications and extraction of the key aspects: Data source, Pre-processing techniques, Feature extraction and embedding techniques, Classification and Performance Evaluation. Section 4 presents the critical review of these key aspects.

\section{Critical Review of the selected studies}

Various attempts have been made by many researchers to create automated systems for assigning clinical codes to patients episode of care \cite{kaur2018}. Research studies have used different methods and techniques to handle and process clinical text, but the standard pipeline is utilised in some shape or form. This section critically reviews the selected studies from six different aspects as mentioned above.

\subsection{Data source} 

Clinical documents can be classified into two categories: clinical notes and diagnostic reports. Clinical notes may include patient's medical history, physical examination history, clinical observations, a summary of diagnostic and therapeutic procedures and treatment plan, whereas diagnostic reports refer to the reports provided by diagnostic services including laboratory reports, radiology reports, and pathology reports. However, in this review, we aim to primarily focus on discharge summaries and other clinical reports used along with discharge summary as the input text data. Table ~\ref{tab:typesDataset} shows a list of datasets used by the selected 38 studies. The datasets: MIMIC-II and MIMIC-III are publicly available, whereas the remaining four, University of Kentucky (UKY) medical centre, Australian hospital medical records, NYU Langone hospital, and Taiwan hospital discharge notes are private datasets obtained from hospitals.

Apart from that, we found that the majority of the studies (n=31) have focused on the ICD-9-CM classification system, 3 studies \cite{Chin2017, Amoia2018, Zachariah2020-bert-xml} predicted ICD-10-CM codes, 2 studies \cite{kaur2018, Rajvir2019} predicted ICD-10-AM and ACHI codes, and 1 study \cite{subotin2014} used ICD-10-PCS, 1 study \cite{Xu-Keyang2019} converted ICD-9-CM codes to ICD-10-CM codes an using online resource before assigning them to discharge summaries. This shows that there is a scarcity of studies relevant to other classification systems including Australian classification systems. 

\begin{table*}[H]

\caption{Types of Datasets}
\label{tab:typesDataset}
\centering
\begin{tabular}{p{3.5cm}p{2cm}p{7.5cm}p{3cm}}
\toprule
 
\textbf{Dataset} & \textbf{Coding system} &  \textbf{Description} & \textbf{Studies} \\ 
\midrule

MIMIC-II & ICD-9-CM & MIMIC-II (Medical Information Mart for Intensive Care II) database contains clinical records for 32,536 subjects collected between 2001 and 2008 from a variety of ICUs (medical, surgical, coronary care, and neonatal) in a single tertiary teaching hospital.  &  \citet{Marafino2014}, \citet{Perotte2013}, \citet{Du2019} \\ \addlinespace[1mm]

MIMIC-III & ICD-9-CM & MIMIC-III database is an extension of MIMIC-II comprising deidentified health-related data associated with over 40,000 patients who stayed in critical care units of the Beth Israel Deaconess Medical Center between 2001 and 2012.  & \citet{Ayyar2016}, \citet{Prakash2017}, \citet{Berndorfer2017}, \citet{Catling2018}, \citet{Samonte2018}, \citet{Xie2018}, \citet{Xie2019}, \citet{Falis2019}, \citet{Huang2019}, \citet{Henning2019}, \citet{Xu-Keyang2019},  \citet{Donglin2020}, \citet{Song2020}, \citet{Teng2020}, \citet{Shaoxiong2020-dilated}, \citet{Chung2020},  \\ \addlinespace[1mm]

MIMIC-II and MIMIC-III & ICD-9-CM & Studies have used both MIMIC-II and MIMIC-III dataset. & \citet{Baumel2018}, \citet{Mullenbach2018}, \citet{rios-kavuluru-2018-shot}, \citet{Min2019}, \citet{Cao2020}, \citet{Thanh2020}, \citet{LiFei2020}  \\ \addlinespace[1mm]

MIMIC and other dataset & ICD-9-CM & Studies have used MIMIC dataset and other sources.  & \citet{Zeng2019}, \citet{Aaron2020}, \citet{Mascio2020-comparative},  \citet{Moons2020}\\ \addlinespace[1mm]

University of Kentucky (UKY) medical centre & ICD-9-CM & The electronic medical records (EMRs) of the UKY medical center in-patient visits with discharge dates in the 2011-2012 two-year period. There are total of 71,461 EMRs having a total of 7,485 unique ICD-9 codes. UKLarge dataset consists of all in-patient visits. A subset of UKLarge, called UKSmall with 1,000 EMRs corresponding to a randomly chosen set of 1,000 in-patient visits from February, 2012.   & \citet{Kavuluru2015}, \citet{Rios2019} \\ \addlinespace[1mm]

Australian hospital medical records & ICD-10-AM and ACHI & A collection of medical records from acute or sub-acute hospitals all over Australia, held by the National Centre for Classification in Health (NCCH).  & \citet{RajvirThesis2018}, \citet{Rajvir2019} \\ \addlinespace[1mm]

Taiwan hospital discharge notes & ICD-10-CM & Discharge notes from The Tri-Service General hospital, Taipei, Taiwan from June 1, 2015 to January 31, 2017  & \citet{Chin2017}  \\ \addlinespace[1mm]

NYU Langone Hospital & ICD-10 & A total of 7.5 million notes corresponding to visits from about 1 million patients. Over 50\% of the notes are progress notes, followed by telephone encounter (10\%) and patient instructions (5\%). & \citet{Zachariah2020-bert-xml} \\ \addlinespace[1mm]

Other healthcare providers & ICD-10-CM, ICD-10-PCS & Studies have not mentioned data sources & \citet{Amoia2018}, \citet{subotin2014} \\

\bottomrule
\end{tabular}
\end{table*}

\subsection{Preprocessing}
  
Preprocessing is done to remove unwanted or meaningless information from the dataset as clinical narratives may contain high level of noise, sparsity, misspelled words, or  grammatical errors. Different preprocessing techniques including tokenization, lowercase conversion, removal of stop words, sentence segmentation, removal of non-alphabetical character, abbreviation expansion, spell error detection and correction, negation detection, stemming, lemmatization, parsing, part-of-speech tagging, named entity recognition, and word normalization were applied in the selected studies as shown in Table~\ref{tab:preprocessingTechniques}. The majority of studies have applied tokenization, followed by removal of stop words, removal of non-alphabetic characters and lowercase conversion. Apart from that, few studies also used other data processing steps such as regular expression matching~\cite{Teng2020}, building dictionary or vocabulary \cite{Baumel2018, Chin2017}, removing non-matching terms \cite{Henning2019}, removal of de-identified or confidential information \cite{Samonte2018, Huang2019}. Studies including \cite{Prakash2017, Baumel2018, Mullenbach2018, Thanh2020, Song2020, LiFei2020, Moons2020, Shaoxiong2020-dilated} truncated the documents to a maximum length of 2,500 or 4,000 tokens in order to reduce the computational cost. Also, \citet{Mullenbach2018, Xie2019, Song2020, Teng2020, Moons2020} replaced tokens with an `UNK' token if they appeared in less than three training documents. Apart from that, studies \cite{subotin2014, Zeng2019, Min2019, Cao2020} may have used any type of data processing technique but have not reported that in their study. 

\clearpage
\onecolumn
\begin{landscape}

\begin{longtable}{p{5cm}p{16cm}p{2cm}}
\caption{NLP techniques for data preprocessing}
\label{tab:preprocessingTechniques}
 \\

\hline  \multicolumn{1}{c}{\textbf{Preprocessing techniques}} &  \textbf{Studies} & \textbf{Study Count}  \\ \hline

\endfirsthead

\multicolumn{3}{c}%
{{\bfseries \tablename\ \thetable{} -- continued from previous page}} \\
\hline 
 \multicolumn{1}{c}{\textbf{Preprocessing techniques}}  & \textbf{Studies}  & \textbf{Study Count}    \\ 

   \hline 
\endhead

\hline \multicolumn{3}{r}{{Continued on next page}} \\ \hline
\endfoot

\hline \hline
\endlastfoot

Tokenization & \citet{Perotte2013}, \citet{Ayyar2016}, \citet{Berndorfer2017}, \citet{Amoia2018}, \citet{Catling2018}, \citet{Baumel2018}, \citet{Mullenbach2018}, \citet{Samonte2018}, \citet{kaur2018}, \citet{Xie2019}, \citet{Falis2019}, \citet{Xu-Keyang2019}, \citet{Du2019}, \citet{Thanh2020}, \citet{Mascio2020-comparative}, \citet{LiFei2020}, \citet{Teng2020}, \citet{Moons2020}, \cite{Shaoxiong2020-dilated} \vspace{1mm} & 19\\

Lowercase conversion & \citet{Amoia2018}, \citet{Mullenbach2018}, \citet{rios-kavuluru-2018-shot}, \citet{Xie2019}, \citet{Falis2019}, \citet{Henning2019}, \citet{Thanh2020}, \citet{Song2020}, \citet{Mascio2020-comparative}, \citet{LiFei2020}, \citet{Teng2020}, \citet{Moons2020}, \cite{Shaoxiong2020-dilated} \vspace{1mm} & 13\\

Removal of stop words & \citet{Marafino2014}, \citet{Kavuluru2015}, \citet{Prakash2017}, \citet{Berndorfer2017}, \citet{Baumel2018}, \citet{Mullenbach2018}, \citet{rios-kavuluru-2018-shot}, \citet{kaur2018}, \citet{Rajvir2019}, \citet{Huang2019}, \citet{Du2019}, \citet{Donglin2020}, \citet{Teng2020}, \citet{Chung2020} \vspace{1mm} & 14 \\

Sentence segmentation or splitting & \citet{Marafino2014}, \citet{Baumel2018}, \citet{kaur2018}, \citet{Chung2020}, \citet{Zachariah2020-bert-xml} \vspace{1mm} & 5 \\

Removal of non-alphabetical characters including punctuation, special characters \vspace{1mm}& \citet{Marafino2014}, \citet{Baumel2018}, \citet{Mullenbach2018}, \citet{Xie2019}, \citet{Falis2019}, \citet{Du2019}, \citet{Thanh2020}, \citet{Song2020}, \citet{Mascio2020-comparative}, \citet{LiFei2020}, \citet{Moons2020}, \cite{Shaoxiong2020-dilated}, \citet{Chung2020} \vspace{1mm} & 13\\

Abbreviation expansion & \citet{kaur2018}, \citet{Rajvir2019} \vspace{1mm} & 2 \\

Spell error detection and correction & \citet{Chin2017}, \citet{kaur2018}, \citet{Rajvir2019} \vspace{1mm}& 3\\

Negation detection & \citet{Marafino2014}, \citet{kaur2018}, \citet{Donglin2020} \vspace{1mm} & 3 \\

Stemming & \citet{Rajvir2019}, \citet{Mascio2020-comparative} \vspace{1mm} & 2 \\

Lemmatization & \citet{Berndorfer2017}, \citet{Mascio2020-comparative} \vspace{1mm} & 2\\

Parsing & \citet{Aaron2020} \vspace{1mm}& 1\\

Part-of-speech tagging & \citet{Berndorfer2017} \vspace{1mm}& 1\\

Named Entity Recognition & \citet{Kavuluru2015}, \citet{Aaron2020} \vspace{1mm}& 2\\

Normalization & \citet{Amoia2018}, \citet{Xu-Keyang2019} \vspace{1mm}& 2 \\

Regular expression matching & \citet{Teng2020} \vspace{1mm} & 1\\ 

Built dictionary or vocabulary & \citet{Chin2017}, \citet{Baumel2018} \vspace{1mm} & 2\\

Truncation of documents & \citet{Prakash2017}, \citet{Baumel2018}, \citet{Mullenbach2018}, \citet{Rios2019}, \citet{Thanh2020}, \citet{Song2020}, \citet{LiFei2020}, \citet{Moons2020}, \citet{Shaoxiong2020-dilated} \vspace{1mm} & 9\\

Replace to `Unkown or UNK' & \citet{Mullenbach2018}, \citet{Xie2019}, \citet{Song2020}, \citet{Teng2020}, \citet{Moons2020} \vspace{1mm} & 5\\

Removal of de-identified or confidential information & \citet{Samonte2018}, \citet{Huang2019} \vspace{1mm} & 2 \\

Extracted diagnosis descriptions & \citet{Xie2018} \vspace{1mm} & 1\\

Removing non-matching terms & \citet{Henning2019} \vspace{1mm} & 1\\

No information & \citet{subotin2014}, \citet{Zeng2019}, \citet{Min2019}, \citet{Cao2020} \vspace{1mm} & 4\\

\end{longtable}

\end{landscape}
\clearpage
\twocolumn

\begin{table*}
\begin{threeparttable}

\caption{Feature Extraction techniques and pre-trained embeddings}
\label{tab:featureTechniques}
\centering

\begin{tabular}{p{0.5cm}p{4cm}p{1.1cm}p{1.2cm}p{0.6cm}p{1cm}p{1.9cm}p{0.7cm}p{3.1cm}}
\hline

\textbf{Year} & \textbf{Study} & \textbf{TF-IDF} &\textbf{N-grams}  & \textbf{BoW} & \textbf{Doc2Vec}  & \textbf{Word2Vec} & \textbf{GloVe} & \textbf{W\textsubscript{embed} dim/ Others} \\ \hline
2013 & \citet{Perotte2013} & \cmark &--  & --  & --  & -- & --& -- \\

2014 & \citet{Marafino2014} & \cmark &\cmark  & --   & -- & -- &-- & -- \\ 

2014 & \citet{subotin2014} &  -- &--  & \cmark & --  & -- & -- & -- \\

2015 & \citet{Kavuluru2015} & \cmark &\cmark  & --  & -- & -- &-- & -- \\

2016 & \citet{Ayyar2016} &-- &--  & -- & -- & -- & \cmark & -- \\

2017 & \citet{Prakash2017} & -- &--  & \cmark   & --  & -- &--& --  \\

2017 & \citet{Chin2017} & -- &--  & --   & --  & -- &\cmark & --  \\

2017 & \citet{Berndorfer2017} & \cmark &--  & \cmark   & --  & \cmark (CBOW) &--  & --  \\

%\cite{Haoran2017} & In Arxiv  \\

%2017 & \cite{SCHEURWEGS2017} & \\

2018 & \citet{Amoia2018} &  -- &--  & \cmark   & --  & \cmark & --  \\

2018 & \citet{Catling2018} & \cmark &--  & --   & --  & \cmark (skip-gram) &-- & --   \\

2018& \citet{Baumel2018} &  \cmark &--  & \cmark  & --  & \cmark (CBOW) & -- & --   \\

2018 & \citet{Mullenbach2018} & -- &--  & \cmark  & --  & \cmark (CBOW) & -- & -- \\

2018 & \citet{Samonte2018}  & -- &--  & --  & --  & -- & -- & \cmark (Topical W\textsubscript{embed})\\

2018 & \citet{Xie2018} &  -- &-- & -- & --  & --& -- & \cmark (200 dim W\textsubscript{embed}) \\

2018 & \citet{rios-kavuluru-2018-shot}&  -- &-- & --  & --  & --& -- & \cmark (300 dim W\textsubscript{embed}) \\

2018 & \citet{kaur2018} & -- &-- &\cmark  & --  & -- & -- & -- \\

%\cite{RajvirThesis2018}  \\

%2018 & \cite{Rajkomar2018scalable} & \\

2019 & \citet{Zeng2019} & -- &--  & --  & --  & -- & -- & \cmark (100 dim W\textsubscript{embed})\\

2019 & \citet{Rajvir2019} & \cmark &--  & --  & --  & -- & -- & -- \\

%2019 & \citet{Gangavarapu2019-TAGS} & -- &--  & -- & \cmark  & \cmark  & -- & -- & --  \\

%\cite{Helwe2017}  \\

2019 & \citet{Xie2019} & -- &--  & --   & --  & \cmark (CBOW) & -- & -- \\ 

2019 & \citet{Falis2019}  & -- &--  & --  & --  & \cmark (CBOW) &-- & --  \\

2019 & \citet{Huang2019}  & \cmark &--  & --  & --  & \cmark (CBOW) & -- & --  \\

2019 & \citet{Min2019} & -- &--  & -- & \cmark  & \cmark (skip-gram) & -- & -- \\

2019 & \citet{Rios2019} & \cmark &--  & --   & --  & -- & -- & \cmark(300 dim W\textsubscript{embed})  \\

2019 & \citet{Henning2019}  &  \cmark &--  & \cmark & --  & -- & -- & -- \\

2019 & \citet{Xu-Keyang2019} &  \cmark &--  & --   & --  & -- & -- & \cmark (256 dim W\textsubscript{embed})\\

%2020 & \cite{Chin2019-Longitudinal} & \\

2019 & \citet{Du2019} &  -- &--  & --  & --  & \cmark & -- & --\\

%2020 & \citet{GANGAVARAPU2020} &  -- &--  & \cmark & \cmark  & \cmark  & -- & -- & -- \\

2020 & \citet{Cao2020}  &   -- &--  & --  & --  & \cmark & -- & --  \\

2020& \citet{Donglin2020}  &  \cmark &--  & --  & --  & \cmark (skip-gram) & -- & -- \\

2020 & \citet{Thanh2020} &  -- &--  & --  & --  & \cmark (CBOW) & -- & -- \\

2020 & \citet{Song2020} &   -- &--  &--  & --  & -- & -- & \cmark (200 dim W\textsubscript{embed})\\

%2020& \citet{Jayasimha2020} &   -- &-- & \cmark & \cmark  & --  & -- & -- & -- \\

2020 & \citet{Aaron2020} & \cmark &-- & --   & --  & -- & -- & \cmark(cui2vec) \\

2020 & \citet{Mascio2020-comparative} & -- &--  & --  & --  & \cmark & \cmark & \cmark (FastText) \\

2020 & \citet{LiFei2020} &  -- &--  & --  & --  & \cmark & -- & --  \\

2020 & \citet{Teng2020} &  -- &--  & --  & --  & -- & -- & \cmark (Knowledge graph)\\

2020 & \citet{Moons2020} &  -- &--  & --  & --  & -- & -- & \cmark(W\textsubscript{embed})\\

2020 & \citet{Shaoxiong2020-dilated} &  -- &--  & --  & --  & \cmark (100 dim) & -- & --\\ 

2020 & \citet{Chung2020} &  \cmark &--  & --  & -- & \cmark & --& -- \\ 
2020 & \citet{Zachariah2020-bert-xml} & -- & -- &-- & -- & -- & -- & \cmark(EHR-BERT)\\
%2020 & \citet{Gangavarapu2020-FarSight} &  -- &--  & \cmark &\cmark  & \cmark & -- & -- & --\\ 

\bottomrule
\end{tabular}

TF-IDF: Term Frequency with inverse Document Frequency; BoW: Bag-of-words; TW: Term Weighting; CBOW: Continuous-bag-of-words; W\textsubscript{embed}: Word embedding; dim: dimension; BERT: Bidirectional Encoder Representations from Transformers

\end{threeparttable}
\end{table*}
\subsection{Feature Extraction}

Feature extraction is the process of extracting useful features from clinical reports. There are two general approaches of feature extraction: expert driven and fully automated feature extraction \cite{Mujtaba2019}. In expert driven feature extraction, a group of experts discovers useful and discriminative features from clinical reports, whereas, the fully automated feature extraction approach makes use of a computer program rather than of any human or expert intervention for extracting the features. 

Different types of feature extraction techniques/word embeddings used in the selected studies listed in Table~\ref{tab:featureTechniques} includes: 

\begin{enumerate}

    \item Term Frequency - Inverse Document Frequency (TF-IDF) evaluates the importance of the word in a document or corpus.
    
    \item Bag-of-Words (BoW): A vocabulary of unique words is created, where each word represents an independent, and discriminative feature.
     
    \item N-gram is the contiguous sequence of words, where N may be `1', `2', '3', and so on. One word sequence is called 1-gram (or unigram), sequence of two words is called 2-gram (or bi-gram), sequence of three words is called 3-gram (Tri-gram), and so on. 
    
    \item Word2Vec is a method to construct embeddings using two models: Skip-gram and Continuous Bag of Words (CBOW). The skip-gram model learns from the existing words available in a sentence to predict the next word, whereas the CBOW model uses the neighboring word to predict the next word \cite{Mujtaba2019}. 
    
    \item Doc2Vec and Paragraph2vec are variants of the word2vec. They focus on predicting words in the document or paragraph. Doc2vec creates a numeric representation of a document irrespective to its length. 
    
    \item Global Vectors (GloVe) is another commonly used word embedding method that derives the relationship between the words from the global corpus statistics. 
    
    \item FastText builds on a specific limitation of word2vec and GloVe. It can handle out-of-vocabulary terms by extending the skip-gram model with internal sub-word information \cite{KHATTAK2019-embedding}.
    
    \item BERT is a contextualised word representation model based on a multi-layer bi-directional transformer-encoder \cite{devlin-2019-BERT}. BERT is pre-trained on two unsupervised tasks: masked language model and next sentence prediction. There are many domain specific versions of BERT available. For clinical text commonly used are:
    \begin{itemize}
        
        \item BioBERT is BERT based model pre-trained on biomedical domain corpora (PubMed abstract and PMC full-text articles).
        
        \item ClinicalBERT is trained on clinical text from MIMIC-III database. \citet{alsentzer2019-ClinicalBERT} presented various versions of BERT  including ClinicalBERT, Clinical BioBERT, Discharge Summary BERT, and Discharge Summary BioBERT.
            
    \end{itemize}
\end{enumerate}

On the other hand, some doctors and physicians make use of different terms (or synonyms) to describe same medical condition in clinical reports. For example, the term `heart attack' or `Myocardial Infarction' belongs to same medical concept. Therefore, for this type of issues, concept-based features are extracted using either medical ontologies such as SNOMED-CT\footnote{https://ontoserver.csiro.au} or some tools such as MetaMap\footnote{https://metamap.nlm.nih.gov}, cTAKES\footnote{https://ctakes.apache.org/}. 

To summarise, the majority of studies (n=17) applied Word2Vec embedding, followed by TF-IDF feature representation and BoW. Few studies \citet{Ayyar2016, Chin2017}, and \citet{Mascio2020-comparative} used GloVe embeddings, while \citet{Samonte2018} applied topical word embedding. There are a few studies that have not reported the embedding model except dimensions of embedding.  \citet{Zachariah2020-bert-xml} trained BERT models from scratch on EHR notes. \citet{Teng2020} applied knowledge graph embeddings to extract entities related to ICD-9 from freebase and their results showed a significant improvement in code prediction.

\subsection{Classification }

In this review, the selected 38 studies have classified the discharge summaries either using machine learning or deep learning models. 

\subsubsection{Traditional Machine Learning (ML) approach}

Machine learning approaches have gained more interest in many clinical research studies due to their efficiency and effectiveness \cite{kaur2018}. Researchers across the globe have employed text classification to categorise clinical narratives into various categories using machine learning approaches including supervised \cite{Hastie2009}, unsupervised \cite{Ko2000}, semi-supervised \cite{zhu2009}, transfer \cite{Sinno}, and multi-view learning approaches \cite{Amini2009}. Among all the aforementioned ML approaches, the supervised ML approach is used often in clinical text classification research. In supervised ML approach \cite{Mujtaba2019}, the clinical narratives collected from hospitals are labeled by domain experts into specific categories. For example, the reports are labeled into two classes, cancer-positive (if the patient is diagnosed with cancer) and cancer-negative (if the patient is not diagnosed with cancer). After labeling, the clinical narratives are pre-processed via NLP techniques so that unnecessary information or noise is removed from the reports. Then, feature engineering is applied to extract the most discriminative features from the clinical narratives and form a numeric feature vector. This feature vector is then provided as an input to ML algorithms such as SVM, Decision Trees, and AdaBoost to construct and validate the classification model. The ML algorithms learn from the data and perform classification, segmentation or prediction. The classification model can further be validated through either random sampling, k-fold cross validation, or leave-one-out techniques \cite{Kohavi1995}. In random sampling, the clinical data are shuffled randomly and divided into training, validation and testing set. For example, 70\% training set, 20\% testing set, and 10\% validation set. Similarly, in k-fold cross validation, the shuffled data are split into data chunks and they perform training on the $(k-1)$ data chunks and test it on the $(k-1)th$ chunk. This process is repeated up to $k$ times. The advantage of using k-fold cross validation method is that all the clinical reports are used for training and testing purposes, and each report is used for testing only once. However, this method is slower than random sampling. Leave-one-out is a special case of the k-fold, where the model is trained with $(n-1)$ reports and tested on the $(n-1)th$ report, leaving one report out each time. However, this method is good when the data are limited and imbalanced, but is much slower than random sampling \cite{Mujtaba2019}.

\subsubsection{Deep Learning}

Deep learning is a type of machine learning technique that utilises a multi-layered neural network architecture to learn the hierarchical representation of data \cite{Hasan2019}. Deep learning models have demonstrated successful results in many NLP tasks such as language translation \cite{machineTranslators}, image captioning \cite{lecun2015} and sentiment analysis \cite{Richard}. The performance of machine learning methods heavily depends on data representation (or features) on which they are applied. Therefore, much of the efforts deploying machine learning algorithms goes into design of pre-processing pipeline and data transformation \cite{Bengio2013}. Deep learning works quite well to solve non-linear classification problems and in the recent years, nonlinear neural network models applied to NLP techniques have achieved promising results over the approaches that use linear models such as SVM and logistic regression \cite{Goldberg2016, Hasan2019}. There are various known neural network models that are used for text and document classification such as Convolutional Neural Networks (CNN), Recurrent Neural Networks (RNN), Long Short Term Memory (LSTM) \cite{Hochreiter1997}, Gated Recurrent Units (GRUs), and Bi-directional Recurrent Neural Networks (BRNN). Though CNN is known for image recognition or visual representation, there are a few studies that have used CNN in sentence classification \cite{Kim2014}. RNN is known for sentence classification with over sequential input and predict sequential output in NLP and other ML tasks. RNN has a bidirectional structure that incorporates both forward and backward inputs, but suffers from vanishing gradient problem. To solve this problem, LSTM is used along with RNN \cite{Samonte2017}.

Over the past two decades, researchers explored various machine learning algorithms to assign ICD codes to patients clinical narratives \cite{Perotte2013, Kavuluru2015, Berndorfer2017, kaur2018, Rajvir2019}. Despite their research efforts, it is believed that the accuracy of assigning clinical codes can be further improved by deep learning approaches. However, the clinical text poses more challenges than general domain text due to various reasons:  

\begin{itemize}
    \item Firstly, the free text clinical narratives contain high levels of noise, sparsity, complex medical vocabulary, misspelled words, abbreviations, use of nonstandard clinical jargons and grammatical errors \cite{Nguyen:2016Noise}.
    \item Secondly, many clinical corpora (or datasets) have imbalanced class distribution. For example, the chances of having cancer positive class are less as compared to cancer negative class. In many cases, a few positive cases in such dataset are likely classified as a rare occurrence, or ignored because it causes more misclassifications compared to the majority class. 
    \item Thirdly, in many clinical narratives, doctors and physicians prefer to use a variety of medical words or phrases. For example, instead of writing myocardial infarction, the phrase heart attack is used. 
    \item Lastly, clinical documents have inconsistent document structure and organisation. Moreover, clinical documents are de-identified and anonymised to ensure patients' data privacy. 
\end{itemize}
   
Table~\ref{tab:classifiersModels} shows machine learning and deep learning models that were employed for assigning ICD codes to discharge summaries.  Notably, in several studies (\citet{Perotte2013, Marafino2014, subotin2014, Ayyar2016, Berndorfer2017, Amoia2018, Catling2018, Baumel2018, kaur2018, Rajvir2019, Xu-Keyang2019, Moons2020}) authors did not compare their proposed model with any existing study or algorithm; therefore, the third column value is left empty. A brief overview and comparison of studies is presented in the Section \ref{discussion}.

\subsection{Evaluation Metrics}

The performance of clinical text classification models can be measured using standard evaluation metrics which include precision, recall, F-measure (or F-score), accuracy, precision (micro and macro-average), recall (micro and macro-average), F-measure (micro and macro-average), and area under the curve (AUC). These metrics can be computed by using values of true positive (TP), false positive (FP), true negative (TN), and false negative (FN) in the standard confusion matrix as shown in Table~\ref{tab:ConfMetric}.

The evaluation metrics for binary classification problems include precision, recall (sensitivity), F-score, accuracy, specificity, and AUC. 

\begin{table}[H]
\centering
\caption{Confusion metric to compare the predicted value with ground truth value}
\label{tab:ConfMetric}
\begin{tabular}{l|l|c|c|c}
\multicolumn{2}{c}{}&\multicolumn{2}{c}{\textbf{Ground Truth}}&\\
\cline{3-4}
\multicolumn{2}{c|}{}&Positive&Negative\\
\cline{2-4}
\multirow{2}{*}& & $True $ & $False$\\
\multirow{2}{*}&Positive  & $Positive $ & $ Negative$\\
\multirow{2}{*}{\textbf{Predicted}}& & $(TP)$ & $(FN)$\\
\cline{2-4}
&  & $False $ & $True $ \\
\multirow{2}{*}&Negative & $Positive$ & $ Negative$\\
\multirow{2}{*}& & $(FP)$ & $(TN)$\\ 
\cline{2-4}

\end{tabular}
\end{table}

\begin{enumerate}
\item \textit{Precision (P)} is the ratio of correct instances retrieved to the total number of retrieved instances and is defined in equation~\ref{precision}. It is also known as positive predictive value (PPV). The ideal value for precision is one (1). The larger the value, the better is the performance.
\begin{equation} \label{precision}
\precision = \frac{\TP}{\TP + \FP}
\end{equation}

\item \textit{Recall (R)} is the number of correct instances retrieved divided by all correct instances and is defined in equation~\ref{recall}. It is also known as sensitivity or true positive rate (TPR). The ideal value for recall is also one (1).
\begin{equation} \label{recall}
\recall = \frac{\TP}{\TP + \FN}
\end{equation}

\item \textit{F-score} is weighted average of precision and recall. High value of F-score represents the best performance, whereas lower value represents the worst performance.

\begin{equation} \label{fscore}
F_1 = \frac{(1+\beta^2)\, \precision \cdot \recall}{\beta^2 \, \precision + \recall}
\end{equation}

when $\beta=1$, then it is called $F_1$ score which is defined as given in equation~\ref{fscore}.

\begin{equation}
F = \frac{2 \, \precision \cdot \recall}{\precision + \recall}
\end{equation}

\item \textit{Accuracy} is the ratio of true instances retrieved to the total number of instances in the dataset and is defined as given in equation~\ref{acc}. 
\begin{equation} \label{acc}
\accy = \frac{\TP + \TN}{\TP+\FP+\FN+\TN}
\end{equation}

\item \textit{Specificity} defines the proportion of negative instances that are correctly predicted as a negative and is formally defined as in equation~\ref{specif}.

\begin{equation} \label{specif}
\spfty = \frac{\TN}{\FP+\TN}
\end{equation}

\item \textit{Area Under the Curve (AUC)} measures the ability of a classifier to distinguish between classes. It plots the rate of true positive (TPR) against the rate of false positive (FPR) \cite{Mujtaba2019}.  A detailed description can be found in \cite{Provost1998, Hand2004, FAWCETT2006}.

\begin{equation}
\text{FPR} = 1- \spfty = \frac{\FP}{\FP+\TN}
\end{equation}

\end{enumerate}

For multi-class problems the performance can be evaluated by measuring micro or macro averaging of precision, recall, F-score, average accuracy, and error rate. Similarly, the performance of multi-label problems can be categorised into three measuring groups: example-based, label-based and ranking-based \cite{Rajvir2019}. The detailed description of other classification problems can be found in \cite{SOKOLOVA2009, GibajaEvaluation2015}.

The macro-averaging and micro-averaging can be calculated as :

\begin{equation*}
B\textsubscript{macro} = \frac{1}{q}  \sum_{i=1}^{q} B(\TP\textsubscript{\textit{i}}, \FP\textsubscript{\textit{i}}, \TN\textsubscript{\textit{i}}, \FN\textsubscript{\textit{i}}) 
\end{equation*}

\begin{equation*}
B\textsubscript{micro} = B(  \sum_{i=1}^{q} \TP\textsubscript{\textit{i}},  \sum_{i=1}^{q} \FP\textsubscript{\textit{i}}, \sum_{i=1}^{q} \TN\textsubscript{\textit{i}}, \sum_{i=1}^{q} \FN\textsubscript{\textit{i}}) 
\end{equation*}

\subsection*{Macro-averaging metric} 
Macro-average metric computes the metric independently for each class and then takes the average. It is basically used when all classes need to be treated equally to evaluate the overall performance of the classifier.

\begin{equation*}
\precision\textsubscript{macro} = \frac{\sum_{i=1}^{q} \TP\textsubscript{\textit{i}}/(\TP\textsubscript{\textit{i}}+\FP\textsubscript{\textit{i}})}{q}
\end{equation*}

\begin{equation*}
\recall\textsubscript{macro} = \frac{\sum_{i=1}^{q} \TP\textsubscript{\textit{i}}/(\TP\textsubscript{\textit{i}}+\FN\textsubscript{\textit{i}})}{q}
\end{equation*}

\begin{equation*}
F_1\textsubscript{macro} =  \frac{2\, \precision\textsubscript{macro} \cdot \recall\textsubscript{macro}}{\precision\textsubscript{macro} + \recall\textsubscript{macro}}
\end{equation*}

\begin{equation*}
\text{Acc} = \frac{\sum_{i=1}^{q} (\TP\textsubscript{\textit{i}}+\TN\textsubscript{\textit{i}})/(\TP\textsubscript{\textit{i}}+\FP\textsubscript{\textit{i}} + \FN\textsubscript{\textit{i}} + \TN\textsubscript{\textit{i}})}{q}
\end{equation*}

\subsection*{Micro-averaging metric}
Micro-average metric aggregates the contribution of individual classes' TP, TN, FP, and FN. It is used to weight each instance or prediction equally.

\begin{equation*}
\precision\textsubscript{micro} = \frac{\sum_{i=1}^{q} \TP\textsubscript{\textit{i}}}{\sum_{i=1}^{q} (\TP\textsubscript{\textit{i}}+\FP\textsubscript{\textit{i}})}
\end{equation*}

\begin{equation*}
\recall\textsubscript{micro} = \frac{\sum_{i=1}^{q} \TP\textsubscript{\textit{i}}}{\sum_{i=1}^{q} (\TP\textsubscript{\textit{i}}+\FN\textsubscript{\textit{i}})}
\end{equation*}

\begin{equation*}
F\textsubscript{micro} =  \frac{2\, \precision\textsubscript{micro} \cdot \recall\textsubscript{micro}}{\precision\textsubscript{micro} + \recall\textsubscript{micro}}
\end{equation*}

Table ~\ref{tab:metrics} shows a list of selected studies that used different evaluation metrics. The evaluation metrics employed in the study is presented with tick [\cmark] mark and the remaining are left empty. To summarise, the majority of studies have evaluated the performance of their model using micro-averaged F1-score, followed by macro-averaged F1-score, and standard F1-score.

\clearpage
\onecolumn
\begin{landscape}

\begin{longtable}{p{0.8cm}p{4.5cm}p{6.5cm}p{10cm}}
\caption{List of machine learning classifiers and deep learning models}
\label{tab:classifiersModels} \\

\hline \textbf{Year} & \textbf{Study}& \multicolumn{1}{c}{\textbf{ML classifiers \& DL models}}  & \multicolumn{1}{c}{\textbf{Compared with other existing studies or algorithms}} \\ \hline 

\endfirsthead

\multicolumn{4}{c}%
{{\bfseries \tablename\ \thetable{} -- continued from previous page}} \\
\hline \textbf{Year} & \textbf{Study} & \multicolumn{1}{c}{\textbf{ML classifiers \& DL models}}  & \multicolumn{1}{c}{\textbf{Compared with other existing studies or algorithms}} \\ \hline 
\endhead

\hline \multicolumn{4}{r}{{Continued on next page}} \\ \hline
\endfoot

\hline \hline 
%\multicolumn{4}{l}{\parbox{\dimexpr\textwidth-2\tabcolsep}{% 
    %\addlinespace[1mm] \footnotesize{ SVM: Support Vector Machine; RF: Random Forest; LR: Logistic Regression; NB: Na\"ive Bayes; MLP: Multi Layer Perceptron; BR: Binary Relevance; ECC: Ensemble of classifier chains; MNB: Multinomial Na\"ive Bayes;  GBM: Gradient Boosting Machine; KV-MemNNs: Key-Value Memory Networks; C-MemNN: Condensed Memory Networks; A-MemNN: Averaged Memory Networks; CNN:  Convolutional Neural Networks; RNN: Recurrent Neural Network; GRU: Gated Recurrent Units; HAN: Hierarchical Attention Network; LSTM: Long Short-Term Memory Networks;  HA-GRU: Hierarchical Attention Gated Recurrent Unit; Conv-LSTM: Convolutional LSTM; EnHANs: Enhanced Hierarchical Attention Networks; CAML: Convolutional Attention for Multi-Label classification; DR-CAML: Description Regularized CAML; DCAN: Dilated Convolutional Attention Network; Seg-GRU: Segment-level GRU; MultiResCNN: Multi-Filter Residual Convolutional Neural Network; C-LSTM-Att: Character-aware LSTM-based Attention; TAGs: Term weighting AGgregated using fuzzy Similarity;  EnTAGs: Enhanced TAGS; ZAGRNN: Zero-shot Attentive Graph Recurrent Neural Networks; ZAGCNN: Zero-shot Attentive Graph Convolution Neural Networks; UNITE: Unsupervised knowledge integration algorithm}}

\endlastfoot

2013 & \citet{Perotte2013} & Flat SVM, Hierarchy-based SVM & --\\

2014 & \citet{Marafino2014} & SVM & -- \\ 

2014 & \citet{subotin2014} & Two-level hierarchical classification & -- \\

2015 & \citet{Kavuluru2015} & SVM, LR, MNB &  BR, copy transformation, ECC  \\

2016 & \citet{Ayyar2016} & LSTM & --\\

%2016 & \citet{Choi2016doctorAI} & \\

2017 & \citet{Prakash2017} & C-MemNN and A-MemNN &  End-to-End Memory Network, KV-MemNNs \\

2017 & \citet{Chin2017} &  CNN & SVM, RF, GBM\\

2017 & \citet{Berndorfer2017} & Flat SVM and Hierarchical SVM & -- \\

%\citet{Haoran2017} & In Arxiv  \\

%2017 & \cite{SCHEURWEGS2017} & \\

2018 & \citet{Amoia2018} & LR and CNN & --\\

2018 & \citet{Catling2018} &  RNN-GRU & -- \\

2018 & \citet{Baumel2018} &  SVM, CBOW, CNN, HA-GRU & -- \\

2018 & \citet{Mullenbach2018} & CAML, DR-CAML  & CNN, LR, Bi-GRU, Flat SVM \cite{Perotte2013}, HA-GRU \cite{Baumel2018}, C-MemNN \cite{Prakash2017}, \cite{Shi2017}, \cite{SCHEURWEGS2017}\\

2018 & \citet{Samonte2018}  & EnHAN & HAN \\

2018 & \citet{Xie2018} & Tree-of-sequences LSTM & HierNet\cite{Yan2015}, HybridNet \cite{Hou2017}, BranchNet \cite{Zhu2017}, LET \cite{Bengio2010}, \cite{Larkey1996}, \cite{Franz2000}, \cite{Kavuluru2013}, \cite{Kavuluru2015}, \cite{Pestian2007} \\

2018 & \citet{rios-kavuluru-2018-shot}& ZACNN, ZAGCNN & LR \cite{Vani2017}, CNN \cite{Baumel2018}, ACNN \cite{Mullenbach2018}, Match-CNN \cite{rios-kavuluru-2018-emr} \\

2018 & \citet{kaur2018} & SVM, NB, Decision Tree, kNN, RF, AdaBoost, and MLP & --
\\

%\cite{RajvirThesis2018}  \\

%2018 & \cite{Rajkomar2018scalable} & \\

2019 & \citet{Zeng2019} & Deep transfer learning using multi-scale CNN and Batch normalisation & Flat SVM \cite{Perotte2013}\\

2019 & \citet{Rajvir2019} & Binary relevance, Label Power set, ML-kNN & --\\

%2019 & \citet{Gangavarapu2019-TAGS} & MLP, LSTM, CNN & \cite{PURUSHOTHAM2018} \\

%\cite{Helwe2017}  \\

2019 & \citet{Xie2019} & MSATT-KG & LR, Selected Feature, Bi-GRU, Flat SVM and Hierarchy SVM, Text-CNN, DR-CAML, CAML, LEAM, C-MemNN, Attentive LSTM \\ 

2019 & \citet{Falis2019}  & Multi-view CNN (Ontological attention ensemble mechanism) & CAML \cite{Mullenbach2018} MVC-LDA and MVC-RLDA \cite{Sadoughi2018} \\

2019 & \citet{Huang2019}  &  CNN, LSTM, GRU & Logistic regression, Random Forest, Feed-forward neural network, C-MemNN \cite{Prakash2017} \\

2019 & \citet{Min2019} & DeepLabeler (CNN and D2V) & Flat SVM and hierarchy-based SVM \cite{Perotte2013} \\

2019 & \citet{Rios2019} & CNNs & Logistic regression, LR+L2R+NERC \cite{Kavuluru2015}, averaging ensemble CNNs without transfer learning \\

2019 & \citet{Henning2019}  & FastText & Binary relevance SVM\cite{Perotte2013}, MT-CNN-net\cite{Du2019}, HA-GRU\cite{Baumel2018}, CAML and DR-CAML\cite{Mullenbach2018}  \\

2019 & \citet{Xu-Keyang2019} & Text-CNN & --  \\

%2020 & \cite{Chin2019-Longitudinal} & \\

2019 & \citet{Du2019} & ML-Net, ML-CNN, ML-HAN  & SVM \cite{Perotte2013}\\

%2020 & \citet{GANGAVARAPU2020} & KNN, MLP, KNN as OvR, LR as OvR, SVM as OvR, RF, HVE, and SE &  \citet{PURUSHOTHAM2018}   \\

2020 & \citet{Cao2020}  & HyperCore &  SVM \cite{Perotte2013}, C-MemNN \cite{Prakash2017}, C-LSTM-ATT \cite{Shi2017}, HA-GRU \cite{Baumel2018}, CAML and DR-CAML \cite{Mullenbach2018}  \\

2020& \citet{Donglin2020}  &  BiLSTMs & DeepLabeler \cite{Min2019} \\

2020 & \citet{Thanh2020} & LAAT and JointLAAT & LR \cite{Mullenbach2018}, SVM \cite{Perotte2013}, CNN \cite{Mullenbach2018}, Bi-GRU \cite{Mullenbach2018}, C-MemNN \cite{Prakash2017}, C-LSTM-Att \cite{Shi2017}, HA-GRU \cite{Baumel2018}, LEAM \cite{Wang2018-joint-embedding}, CAML \cite{Mullenbach2018}, DR-CAML \cite{Mullenbach2018}, MSATT-KG \cite{Xie2019}, MultiResCNN \cite{LiFei2020}\\

2020 & \citet{Song2020} & ZAGRNN, ZAGRNN with LDAM loss & ZAGCNN \cite{rios-kavuluru-2018-shot}, \cite{Liu2019CVPR}, \cite{Xian2018-zeroshot}, \cite{Felix2018} \\

%2020& \citet{Jayasimha2020} & EnTAGs approach using CNN, LSTM, CNN-LSTM, and GRU & TAGS model \cite{Gangavarapu2019-TAGS}  \\

2020 & \citet{Aaron2020} & UNITE & LR, Topic modeling, MLP\\

2020 & \citet{Mascio2020-comparative} & ANN, CNN, Bi-LSTM & SVM \\

2020 & \citet{LiFei2020} &  MultiResCNN & SVM (Flat and Hierarchy) \cite{Perotte2013}, CAML \& DR-CAML \cite{Mullenbach2018}, HA-GRU \cite{Baumel2018}, C-MemNN \cite{Prakash2017}, C-LSTM-Att \cite{Shi2017}\\

2020 & \citet{Teng2020} & G\textunderscore Coder (Multi-CNN, Graph Presentation, Attention Matching, Adversarial Learning) & C-LSTM-Att \cite{Shi2017}, CAML \& DR-CAML \cite{Mullenbach2018}, MultiResCNN \cite{LiFei2020}  \\

2020 & \citet{Moons2020} & CNN, Bi-GRU, DR-CAML, MVC-LDA, MVC-RLDA & --   \\

2020 & \citet{Shaoxiong2020-dilated} & DCAN & CNN \cite{Kim2014}, C-MemNN \cite{Prakash2017}, Attentive LSTM \cite{Shi2017}, Bi-GRU \cite{Mullenbach2018}, CAML \& DR-CAML \cite{Mullenbach2018}, LEAM \cite{Wang2018-joint-embedding}, MultiResCNN \cite{LiFei2020}  \\ 

2020 & \citet{Chung2020} &  CNN, LSTM, GRU, HAN & SVM\\ 

2020 & \citet{Zachariah2020-bert-xml} & AttentionXML (BERT-XML) & LR, Multi-head Attention , BERT, BioBERT, ClinicalBERT.

\end{longtable}
SVM: Support Vector Machine; RF: Random Forest; LR: Logistic Regression; NB: Na\"ive Bayes; MLP: Multi Layer Perceptron; BR: Binary Relevance; ECC: Ensemble of classifier chains; MNB: Multinomial Na\"ive Bayes;  GBM: Gradient Boosting Machine; KV-MemNNs: Key-Value Memory Networks; C-MemNN: Condensed Memory Networks; A-MemNN: Averaged Memory Networks; CNN:  Convolutional Neural Networks; RNN: Recurrent Neural Network; GRU: Gated Recurrent Units; HAN: Hierarchical Attention Network; LSTM: Long Short-Term Memory Networks;  HA-GRU: Hierarchical Attention Gated Recurrent Unit; Conv-LSTM: Convolutional LSTM; EnHANs: Enhanced Hierarchical Attention Networks; CAML: Convolutional Attention for Multi-Label classification; DR-CAML: Description Regularized CAML; DCAN: Dilated Convolutional Attention Network; Seg-GRU: Segment-level GRU; MultiResCNN: Multi-Filter Residual Convolutional Neural Network; C-LSTM-Att: Character-aware LSTM-based Attention; TAGs: Term weighting AGgregated using fuzzy Similarity;  EnTAGs: Enhanced TAGS; ZAGRNN: Zero-shot Attentive Graph Recurrent Neural Networks; ZAGCNN: Zero-shot Attentive Graph Convolution Neural Networks; UNITE: Unsupervised knowledge integration algorithm
\end{landscape}
\clearpage
\twocolumn

\begin{landscape}
\begin{table}
\caption{Studies using different Evaluation Metrics}
\label{tab:metrics}
\centering
\begin{tabular}{ p{0.6cm} p{0.6cm} p{0.5cm} p{0.5cm} p{0.6cm} p{0.5cm} p{0.7cm} p{0.4cm} p{0.4cm} p{0.4cm} p{0.4cm} p{0.4cm} p{0.8cm} p{0.5cm} p{0.5cm} p{0.5cm} p{0.8cm} p{0.9cm} p{0.9cm}  p{0.5cm} p{0.5cm} p{0.5cm} p{0.5cm} p{0.5cm}} 
  \toprule
 \multirow{2}{*}{Year} & \multirow{2}{*}{Study}   & \multicolumn{22}{c}{Evaluation Metrics} \\ \cmidrule{3-24}

   & &  P & R & F1 & Acc & AUC & HL & JS & P\textsubscript{mi} & R\textsubscript{mi} & F1\textsubscript{mi} & AUC\textsubscript{mi} & P\textsubscript{ma} & R\textsubscript{ma} &  F1\textsubscript{ma} & AUC\textsubscript{ma} & F1\textsubscript{diag} & F1\textsubscript{proc}  & P@n & R@n & Sen &Spec & MRR\\
   \midrule 
   
 2013 & \cite{Perotte2013} &  \cmark & \cmark & \cmark & -- & -- &  -- & -- &  --&  --& -- & -- & -- & -- &  -- & -- & -- & --  & -- & -- & -- & -- & --\\
   
 2014 & \cite{Marafino2014} & \cmark & \cmark & \cmark & \cmark & -- & -- & --& -- & -- & -- & -- & -- & -- & --  & -- & -- & -- &-- & -- & -- &-- & --\\

 2014 & \cite{subotin2014} & \cmark & \cmark & \cmark & -- & -- & -- & -- & -- & -- & -- & -- & -- & -- &  -- & -- & -- & -- & -- & -- & -- &-- & \cmark \\

 2015 & \cite{Kavuluru2015} &  -- & -- & -- & -- & --   & -- & -- & \cmark & \cmark  & \cmark  & -- & -- & -- &  -- & -- &-- &--  & -- & -- & -- & -- & --\\
 
 2016 & \cite{Ayyar2016} & \cmark & \cmark & \cmark & -- & --   & -- & -- & -- & -- & -- & -- & -- & -- &  -- & -- & -- & --  & -- & -- & -- &-- & --\\

2017 & \cite{Prakash2017} &  -- & -- & -- & -- & -- & \cmark & -- & -- & -- & -- & -- & -- &  --& --  & \cmark & -- & --  & \cmark & -- & -- &-- &-- \\

2017 & \cite{Chin2017} &  -- & -- & \cmark & -- & --   & -- & -- & -- & -- & -- & -- &--  &  --& --  & -- &  --&  --  & -- & -- & -- & -- &--\\
  
 2017 & \cite{Berndorfer2017} &  \cmark & \cmark & \cmark & -- & --  & -- & -- &--  &--  & -- & -- &--  & -- & --  & -- & -- & --   & -- & -- & -- & -- &--\\
 
 2018 & \cite{Amoia2018} & --  & -- & -- &--  &--  & --  & -- & \cmark & \cmark & \cmark & -- & -- & -- & --  & -- & -- & --   & --&-- & --& -- &--\\
 
 2018 & \cite{Catling2018} &  \cmark & \cmark & \cmark & -- &  --  & -- & -- &--  & -- & -- & -- &--  & -- & --  &--  & -- & --   & -- &--  &--  & --&--\\
   
 2018 & \cite{Baumel2018} &  -- &-- & -- & -- &  -- & -- & -- & -- & -- & \cmark & -- & -- &  --&  -- & -- &--  &  --  & -- &--  &--  & -- &--\\
   
 2018 & \cite{Mullenbach2018} & --  & --& -- & -- & --  & -- & --  &--  & -- & \cmark  & \cmark & -- & -- &  \cmark & \cmark & \cmark & \cmark   & \cmark & --& --&-- &-- \\

 2018 & \cite{Samonte2018} &  \cmark & \cmark & \cmark & \cmark & -- & -- &-- &--  & --  & -- & -- & -- & -- & --  & -- & -- & --  & -- & -- & --& -- &--\\
   
 2018 & \cite{Xie2018} & -- & -- & -- & -- & --  & -- & -- & -- & -- & -- & -- & -- & -- &  -- & -- & -- & --  & -- & -- & \cmark & \cmark &--\\  
   
 2018 & \cite{rios-kavuluru-2018-shot}& --  & -- & -- & -- & -- & -- & -- & -- & -- &--  &  --& -- & -- & --  &--  & -- & --  & -- & \cmark & --& -- &--\\
   
 2018 & \cite{kaur2018} &  \cmark & \cmark & \cmark & \cmark & --  & \cmark & \cmark &--  & -- & -- & -- & -- & -- & -- & -- & -- & --  & -- &  --&  --& -- & --\\
  
 2019 & \cite{Zeng2019} &  -- & -- & -- & -- & -- & -- & -- & \cmark & \cmark  & \cmark  & -- & -- & -- & --  & -- &-- &-- &  --& -- &  --& -- &--\\
  
 2019 &\cite{Rajvir2019}  &  -- & -- & -- & -- & -- & \cmark  & \cmark  & \cmark  & \cmark  & \cmark  &  --& \cmark & \cmark  &  \cmark  & -- & -- & --&--  & -- & -- & -- &--\\

%2019 & \cite{Gangavarapu2019-TAGS} & --  & -- & \cmark & \cmark & \cmark & \cmark  & -- & --  & -- & -- & -- & -- & -- & -- & --  & -- & -- & -- & \cmark & \cmark& \cmark & -- & -- & -- & --\\
 
2019 & \cite{Xie2019} & -- & -- & -- & -- & --  & -- & -- & -- & -- & \cmark & \cmark & -- & -- &  \cmark & \cmark & \cmark & \cmark & \cmark & -- & -- & -- &-- \\

2019 & \cite{Falis2019} & --  & -- & -- & -- & --  & -- & -- &  \cmark &  \cmark &  \cmark &  --&  \cmark &  \cmark &   \cmark & -- & -- & -- &  \cmark & --& -- & --&--\\

 2019 & \cite{Huang2019} &  \cmark & \cmark & \cmark & \cmark & --  & \cmark & -- & -- & -- & -- & -- & -- & -- &  -- & \cmark & -- & -- & \cmark & -- & -- &-- &--\\ 
 
 2019 &  \cite{Min2019}  & --  & -- & -- & -- & --  & -- & -- & \cmark & \cmark & \cmark & -- & -- & -- & --  & -- & -- & --  & -- & -- &-- & -- &--\\
 
 2019 & \cite{Rios2019} & --  & --  & --  & --  & --    & --  & --  & --  & --  & \cmark & --  & --  & --  &  \cmark & --  & --  & --   & --  & --  & --  &-- & --\\ 
 
 2019 & \cite{Henning2019} & \cmark & \cmark & \cmark &  --  &  --   &  --  &  --  &  --  &  --  &  --  &  -- &  --  &  --  &   --  &  --  &  -- &  --   & \cmark &  --  &  --  & --   &--\\ 
 
 2019 & \cite{Xu-Keyang2019} & -- & -- & -- & -- & -- & -- & \cmark & -- & -- & \cmark & \cmark & -- & -- &  \cmark & \cmark & -- & -- & -- & -- & -- &-- &-- \\
 
 2019 & \cite{Du2019} & \cmark & \cmark & \cmark & -- & --  & -- & -- & -- & -- & -- & -- & --& -- &  -- & -- & -- & --  & -- & -- & -- &--&--\\
 
% 2020 & \cite{GANGAVARAPU2020} & --  & -- & \cmark & \cmark & \cmark & \cmark  & -- & --  & -- & -- & -- & -- & -- & -- & --  & -- & -- & -- & \cmark & \cmark& \cmark & -- & -- & -- & --\\
 
  2020 & \cite{Cao2020} & --  & -- & -- & -- & --  & --  & -- & -- & -- & \cmark & \cmark & -- & -- &  \cmark & \cmark & -- & --  & \cmark & --  & -- & -- &-- \\
 
 2020 & \cite{Donglin2020} &  --  &  --  &  --  &  --  & --  & -- & -- & \cmark & \cmark & \cmark & \cmark & \cmark & \cmark &  \cmark & \cmark & -- & --  & -- & -- & -- &-- &-- \\

 2020 & \cite{Thanh2020} &  -- & -- & -- & -- & --  & -- & -- & -- & -- & \cmark & \cmark & -- & -- &  \cmark & \cmark & -- & --  & \cmark& -- & --& -- &--\\

  2020 & \cite{Song2020} & --  & -- & -- & -- & --  & -- & -- & \cmark & \cmark & \cmark & \cmark & \cmark & \cmark &  \cmark & \cmark & -- & --   & -- & -- & -- & --&--\\
  
% 2020 & \cite{Jayasimha2020} & --  & -- & \cmark & \cmark & \cmark & \cmark  & -- & -- & -- &  --& -- & -- & -- & -- & --  & -- & -- & -- & \cmark & \cmark& \cmark & -- & -- & -- & -- \\
 
 2020 & \cite{Aaron2020} &  -- & -- & \cmark & -- & \cmark  & -- & -- & -- & -- & -- & -- & -- & -- &  -- & -- & -- & --  & -- & -- & -- &-- &--\\

 2020 & \cite{Mascio2020-comparative} & -- & -- & \cmark (Avg) & -- & --  & -- & -- & -- & -- & -- & -- & -- & -- &  -- & -- & --& --  & -- & -- & -- &--&--  \\
   
2020 & \cite{LiFei2020} &  -- & -- & -- & -- & --  & -- & -- & -- & -- & \cmark & \cmark & -- & -- &  \cmark & \cmark & -- &--  & \cmark & -- &  --& -- &--\\

2020 & \cite{Teng2020} & -- & -- & -- & -- & -- & -- & -- & -- & -- & \cmark & \cmark & -- & -- &  -- & -- & -- & --  & \cmark & -- & -- &-- &-- \\ 

2020 & \cite{Moons2020} &-- & -- & -- & -- & --  & -- & -- & -- & -- & \cmark & \cmark & -- & -- &  \cmark & -- & \cmark (micro) & \cmark (micro)  & \cmark & -- & -- &--&-- \\

2020 & \cite{Shaoxiong2020-dilated} & -- & -- & -- & -- & -- & -- & -- & -- & -- & \cmark & \cmark & -- & -- &  \cmark & \cmark & -- & -- & \cmark & -- & -- &-- &--\\ 

2020 & \cite{Chung2020} & -- & -- & -- & -- & --  & -- & -- & -- & -- & \cmark & -- & -- & -- &  -- & -- & -- & --  & -- & -- & -- &-- &--\\ 

%2020 & \cite{Gangavarapu2020-FarSight} & -- & -- & \cmark & \cmark & \cmark & \cmark  & -- & -- & -- & -- & -- & -- & -- & -- &  -- & -- & -- & -- & \cmark & --& -- & -- & -- & -- &-- \\
 
2020 & \cite{Zachariah2020-bert-xml} & -- & -- & -- & -- & -- & -- & -- & -- & -- & -- & \cmark & -- & -- &  -- & \cmark & -- & --  & -- & -- & -- &-- & --\\

% & \cite{Choi2016doctorAI} & \\
% & \cite{Shickel2018} & \\
 
% & \cite{Rajkomar2018scalable} & \\
 \bottomrule
\end{tabular}
\\
 \footnotesize{P: Precision, R: Recall, F1: F1-score, Acc: Accuracy, AUC: Area under the ROC curve, HL: Hamming Loss, JS: Jaccard Similarity,  P\textsubscript{mi}: Micro-averaged Precision,  R\textsubscript{mi}: Micro-averaged Recall, F1\textsubscript{mi} or F-meas\textsubscript{mi}: Micro-averaged F-measure, AUC\textsubscript{mi}: Micro-averaged F1 score of Area under the ROC curve (AUC), P\textsubscript{ma}: Macro-averaged Precision,  R\textsubscript{ma}: Macro-averaged Recall, F1\textsubscript{ma}: Micro-averaged F1-score, AUC\textsubscript{ma}: Macro-averaged F1 score of Area under the ROC curve (AUC), F1\textsubscript{diag}: Micro- F1 on diagnosis codes,  F1\textsubscript{proc}: Micro-F1 on procedure codes, P@n: Precision @ n, R@n: Recall @ n, Sen: Sensitivity, Spec: Specificity, MRR: Mean Reciprocal Rank}

\end{table}
\end{landscape}

%\subsection{Results of Phase 1}
%Number of publications in each database\\
%Number of publications after reviewing titles and abstract
%\subsection{Results of Phase 2}
%Number of publications after reviewing full text pdf.

\section{Discussion} \label{discussion}

%Automated ICD coding has been an active research area in healthcare sector from more than two decade \cite{Thanh2020}. 

With the rise of EHR implementation, applications of machine learning and deep learning models to predict clinical events and outcomes for clinical decision making have sparked widespread interest. Due to high granularity of ICD codes, researchers performed clinical code prediction either based on category-level or full-code prediction. The category-level prediction can also be referred to as group-level or chapter-level prediction in which a set of similar diseases and other underline health conditions are represented in a unique chapter or a group. This type of problem is a multi-label classification problem where clinical reports can be mapped to more than one group.  On the other hand, full code prediction is associated with a five characters (numeric or alphanumeric) code where, the first three characters specify the disease category, while the latter two characters provide a more meticulous  division of the disease \cite{Donglin2020}. Furthermore, a few research studies \cite{Donglin2020, Moons2020} focused on the first three characters of the code, which is similar to category-level prediction and treats the task as multi-label classification problem. 

Research studies related to automated ICD coding have used different methods and techniques ranging from pattern matching to deep learning approaches to categorise clinical narratives into different categories. In this review, we are considering the research studies associated with automated ICD coding task that got published from January 1, 2010 on wards; therefore, this systematic literature review is limited to only machine learning and deep learning methods. The studies published prior to the year 2010 are basically focused on pattern matching, rule-based, machine learning or hybrid methods.

\subsection{Traditional Machine Learning (ML)}

In 2013, \citet{Perotte2013} explored the traditional machine learning algorithm to build  \textit{flat} and \textit{hierarchical} SVM for automatically assigning ICD-9 codes to discharge summaries. Their results indicate that the hierarchical SVM outperforms the flat SVM and achieve F-measure 39.5\% and 27.6\% respectively.  \citet{Berndorfer2017} also used predictive models, flat and hierarchical SVM, for ICD-9 coding of discharge summaries using shallow and deep text representation. The results showed that the union prediction model outperformed the BoW representation in terms of F-score by 2.21 points in the flat SVM model and 1.28 points in the hierarchical setting. Similarly, \citet{Marafino2014} developed SVM-based classifiers using n-gram feature extraction that can accurately identify a range of procedures and diagnoses in ICU clinical notes. Their SVM algorithm distinguishes itself on the basis of generalizability, modularity and scalability that make it more accurate than other existing machine learning methods when applied to datasets with multiple variables. However, the study was incorporated with several limitations. (1) Only certain diagnoses and procedures were considered more precisely, 2 diagnosis (Jaundice and Intracranial Haemorrhage (ICH)) and 2 procedures (Ventilation and Phototherapy). (2) The comparison of their approach against gold standards was imperfect, which creates the possibility of incorrect measurement of accuracy. The results show that among all four classification tasks, ventilation classification task outperformed and achieved accuracy of 0.982 and F1-score of 0.954.

\citet{Kavuluru2015} performed empirical evaluation of supervised learning approaches, such as SVM, Na\"ive Bayes, and logistic regression, to assign ICD-9-CM codes to clinical narratives of patients discharged from the UKY medical center. In addition, they also used two other smaller datasets- UKSmall (a subset of UKLarge), and CMC. 
The results were evaluated using multiple complementary methods including example-based measures, and macro and micro-averaging measures. \citet{kaur2018} performed a comparative analysis of seven traditional machine learning algorithms, such as SVM, Na\"ive Bayes, Decision Tree, kNN, Random Forest, AdaBoost, and MLP,  for assigning ICD-10-AM and ACHI codes to discharge summaries acquired from Australian hospital. Among all the classifiers, AdaBoost outperforms by achieving F-score of 0.9141, accuracy of 0.8611, Jaccard similarity of 0.8294 and Hamming Loss of 0.0945 for 235 medical records. Similarly, for 190 records, Decision Tree outperforms and achieves F1-score of 0.8730, accuracy of 0.7920, Jaccard similarity of 0.7453, and Hamming Loss of 0.0877. A year later, the same set of authors in the study \cite{Rajvir2019} aimed to classify multi-label diseases and interventions of the respiratory and digestive system using transformation method (Binary Relevance and Power Label set) and adaptive algorithm (ML-kNN). However, due to limited number of medical records, the dataset was repeated with some minor changes in order to train the model. Also, more than half of the clinical codes appeared only once in the whole dataset, which lowered the learning rate of the system.

\subsection{Deep Learning}

\citet{Ayyar2016} proposed a deep learning framework using LSTM model to classify the ICD-9 codes to patients' discharge summary. They mapped all codes to top level representation, which left top 19 level ICD-codes. The study achieved precision of 0.799, recall of 0.685 and F1-score of 0.708 when used note length of 1000 words. However, the study found a few limitations such as several misspelling occurrences in the data, model not being capable to learn and hold important words or text  in memory for longer period of time, and learning task possible to be improved by use of medical dictionaries so that more pertinent word vectors are obtained. 

\citet{Baumel2018} proposed a Hierarchical Attention-Gated Recurrent Unit (HA-GRU) with two levels of bidirectional GRU for sentence and document encoding to assign multiple ICD codes to discharge summaries of the MIMIC-II and MIMIC-III dataset. The study found that HA-GRU gave better results in the rolled-up ICD-9 setting over the CNN and SVM and evaluated the performance of the model by calculating F1\textsubscript{micro} of 40.52\% on full code set and 53.86\% on rolled up ICD-9 codes (3-digit codes). Later on, \citet{Samonte2018} developed Enhanced hierarchical attention networks (EnHANs) model, which is a modified version of HAN model, to assign ICD-9 codes to patient notes. Patient records with the word ``Discharge Summary'' were processed and mapped only top 10 ICD-9 codes. The study modified the architecture of hierarchical attention networks by adding topical word embedding and a word input.  The experiments were conducted for the 19 label and 10 label tasks. To achieve the best model, only 100 sentences per document, 20 words per sentence were considered and sigmoid binary cross entropy was applied as the loss function to improve the model's performance. While experimenting on 19 label task, the need for regularization was observed when overfitting occurred around 11th-15th epoch. For the validation of results, accuracy, precision, recall and F1-score were evaluated. However, we have recalculated the F1-score as the information provided in the results and analysis section did not match the performance table of the 19 label task. The study also measured the performance of the baseline model HAN for 10 label task and achieved 83\% accuracy, 76.5\% precision, 62.6\% recall and 68.2\% F1-score.

\citet{Mullenbach2018} propose an attention based CNN model which is the combination of a single filter CNN and label dependent attention  for ICD coding of discharge summaries. They call their method Convolutional Attention for Multi-Label classification (CAML) while its variant is referred to as Description Regularized-CAML (DR-CAML). As the model contains only one convolutional filter, it may not be sufficient to learn decent document representations from a flat and fixed-length convolutional architecture \cite{LiFei2020}. Their results show that DR-CAML performs slightly better than CAML and achieves 0.457 F1\textsubscript{micro}  and 0.049 F1\textsubscript{macro} on MIMIC-II dataset. The study outperforms the existing state-of-the-art models \cite{Perotte2013, Baumel2018} on MIMIC-II. On the other hand, CAML outperforms DR-CAML on MIMIC-III and achieves F1\textsubscript{micro} of 0.539, F1\textsubscript{macro} of 0.088, F1\textsubscript{Diag} of 0.524, F1\textsubscript{Proc} of 0.609, P@8 of 0.709 and P@15 of 0.561. A study by \citet{Xie2018} propose a tree-of sequences LSTM architecture to capture the hierarchical relationship among codes and the semantics of each code. In addition, the study also used adversarial learning approach to reconcile the heterogeneous writing style. It also uses attention mechanism to perform many-to-one and one-to-many mappings between diagnosis descriptions and codes. The model outperforms the other baseline models\cite{Larkey1996, Bengio2010, Franz2000, Kavuluru2013, Kavuluru2015, Hou2017, Pestian2007, Yan2015, Zhu2017} by achieving 0.29 sensitivity and 0.33 specificity. However, there are two major limitations found in this study. (1) It does not perform well on infrequent codes, and (2) it is less capable to deal with abbreviations. In the same year, \citet{Amoia2018} address data imbalance issues by implementing two system architectures using convolutional neural networks and logistic regression model to predict ICD-10-CM codes for data acquired from 10 healthcare providers which covers 17 months of data. The study has found that the combination of LR-CNN gives better performance with micro-F1 score of 64.60\% on all seen codes. Furthermore, \citet{Catling2018} have explored different methods for representing clinical text and the labels in hierarchical clinical coding ontologies. For atomic representations of 17,561 disease labels, the weighted F1 increased to 0.264-0.281 from 0.232-0.249. Similarly, RNN text representation improved weighted F1 for the prediction of 19 disease category labels to 0.682-0.701 from 0.662-0.682 using TF-IDF. 

\citet{rios-kavuluru-2018-shot} have performed a fine-grained evaluation of rare and zero-shot label learning using neural network models (ZACNN and ZAGCNN) which utilise ICD hierarchy information for improving the performance on the rare and zero-shot codes. However, the model hardly assigns rare codes in its final prediction \cite{Song2020}. For the frequent and few-shot labels, the study compares their models with the existing models: Attention based CNN\cite{Mullenbach2018}, CNN\cite{Baumel2018}, L1 regularized logistic regression model \cite{Vani2017} and Match-CNN \cite{rios-kavuluru-2018-emr}. The results of this study are given in Table~\ref{tab:comparisonOfStudies1}. Similarly, in 2019 the same two authors studied \cite{Rios2019} the effect of transfer learning for ICD code prediction on 71,463 EMRs from the UKY hospital. The study also introduced a simple transfer learning method that improves prior transfer learning approaches and their prior methods for the UKLarge dataset \cite{Kavuluru2015}. However, the study claims a major weakness which is similar to the other transfer learning methods- training the model on two different datasets which is an acceptable weakness because only training time is increased.

\citet{Henning2019} introduced SVM as the baseline and FastText as the main model with UMLS mapping into word embedding models to assign ICD codes to discharge summaries extracted from MIMIC-III database. The model achieved F1-score of 62.2\% and outperformed previous models \cite{Du2019, Mullenbach2018, Perotte2013}. Following the same year, another study \cite{Min2019} proposed a deep learning framework called DeepLabeler that combines CNN with document to vector (D2V) techniques to automatically assign ICD codes. The study performed experiments on MIMIC-II and MIMIC-III dataset and demonstrated that DeepLabeler outperformed the hierarchy-based SVM with micro F-measure of 0.335 on MIMIC-II and 0.408 on MIMIC-III data. However, the study found two major defects when used CNN to extract local features. 1) It ignores the semantic features of the full text because it does not take into account the order or words or phrases. 2) CNN model takes only the unchanged shape matrix as an input which means that document matrix is either truncated or padded by some zeros, causing certain proportion of the original information to become lost. Therefore, the study has integrated the CNN and D2V parts to achieve the better performance. \citet{Xie2019} improved the convolutional attention model \cite{Mullenbach2018} by using densely connected CNN and multi-scale feature attention on MIMIC-III dataset. This study used a graph convolutional neural network to capture the hierarchical relationships among medical codes and the semantics of each code.

\citet{Xu-Keyang2019} built a hybrid system that includes the CNN, LSTM and decision tree to predict ICD diagnostic codes from different modalities including unstructured, semi-structured and structured tabular data. They performed one-to-one mapping of 32 ICD-9 codes to ICD-10 codes using an online resource~\footnote{https://www.icd10data.com/} and developed an ensemble-based approach which integrated modality-specific models for improving prediction accuracy by achieving F1\textsubscript{micro} of 0.7633 and micro-AUC of 0.9541. However, there is plenty of room for improvement as the study has focused on only 32 ICD-10 codes, and used large feature dimensions and some features are duplicate.

\citet{Huang2019} performed an empirical evaluation of deep learning models for automatic ICD-9 code assignment from the MIMIC-III discharge summaries. The findings showed that deep learning methods (CNN, LSTM, GRU) outperformed the traditional machine learning methods (Logistic Regression, Random Forest, Feed Forward Neural Networks) for predicting the top 10 ICD-9 codes and categories with highest F1 of 0.6957 (GRUs) and 0.5320 (Logistic Regression) for top 10 ICD-9 codes, and with F1-scores of 0.7233(GRUs) and 0.6313 (FNNs) for top 10 categories. The results also showed that the top 50 ICD-9 codes and categories did not outperform the baseline because the model could not effectively distinguish between 50 different labels. In addition, the samples for labels 11 to 50 were highly imbalanced; due to this, the deep neural network could not learn adequately useful representation. \citet{Falis2019} structured an ontological attention ensemble mechanism that matched the structure of the ICD ontology; in its shared attention vectors learnt at each level of the hierarchy are combined into label-dependent ensembles. The study compared the proposed model with the previous state-of-the-art models \cite{Mullenbach2018, Sadoughi2018} and achieved highest F1\textsubscript{micro} of 0.560 and highest p@8 score of 0.727. In addition, the study also highlighted the limitations found in MIMIC-III dataset, in particular the labelling inconsistencies that aroused due to variable coding practices between clinical coders and the need for revision of coding standards.

A year later, in 2020, \citet{LiFei2020} proposed a multi-filter residual CNN (MultiResCNN) for ICD coding of discharge summaries. The MultiResCNN combines a multi-filter convolutional layer and a residual convolutional layer to improve the convolutional attention model proposed by Mullenbach \cite{Mullenbach2018}. The study compared the MultiResCNN with 5 existing models which include SVM \cite{Perotte2013}, HA-GRU \cite{Baumel2018}, CAML \& DR-CAML \cite{Mullenbach2018}, C-MemNN \cite{Prakash2017}  and C-LSTM-Att \cite{Shi2017}. On MIMIC-III full codes, MultiResCNN outperformed CAML \& DR-CAML model in Macro-AUC, F1\textsubscript{micro}, P@k. Similarly, the MultiResCNN also outperformed in all evaluation metrics on MIMIC-III 50 and MIMIC-II. Similarly, \citet{Cao2020} propose a hyperbolic and co-graph representation framework for automatic ICD coding task, which can jointly exploit code hierarchy and code co-occurrence. Specifically, a hyperbolic representation method leverages the code hierarchy in the hyperbolic space. The study also uses the graph convolutional network (GCN) to capture the co-occurrence correlation. The study compares their model with the existing baseline models and outperformes in all evaluation metrics on MIMIC-II, MIMIC-III full, and MIMIC-III 50 dataset as shown in Table\ref{tab:comparisonOfStudies1}, Table\ref{tab:comparisonOfStudies2}, and Table~\ref{tab:comparisonOfStudies3} respectively. To check the effectiveness of the hyperbolic and co-graph representation, the study has also conducted the ablation where HyperCore improves the micro-F1 score from 0.539 to 0.551 on MIMIC-III full data, and causes a 2.6\% improvement of F1\textsubscript{micro} on MIMIC-III 50 dataset. The study also found that when the hyperbolic and co-graph representation is removed, F1\textsubscript{micro} drops from 0.477 to 0.439 on MIMIC-II dataset, which indicates the effectiveness of hyperbolic and co-graph representation. 

\citet{Thanh2020} proposed a new label attention model, LAAT for ICD coding, which can handle the various lengths as well as the interdependence between text fragments related to ICD codes. This study also proposed a hierarchical joint learning mechanism (JointLAAT) to handle imbalanced data problems using the hierarchical structure of the ICD codes. The study compared their proposed model to baseline models including conventional machine learning \cite{Perotte2013} and deep learning models \cite{Mullenbach2018, Prakash2017, Shi2017, Baumel2018, Wang2018-joint-embedding, Xie2019, LiFei2020} on ICD coding and used both MIMIC-II and MIMIC-III dataset. Their results on MIMIC-III have shown that LAAT gives higher results in micro-AUC, F1\textsubscript{micro}, F1\textsubscript{macro}, P@8 and P@15 compared to MASATT-KG \cite{Xie2019} and MultiResCNN \cite{LiFei2020}. The results become improved in macro-AUC and F1\textsubscript{ma} when JointLAAT is used. On the other hand, LAAT outperforms all the baseline models across all metrics on MIMIC-III 50 (when only top 50 codes are considered). There is no significant difference in the results of LAAT and JointLAAT because of no infrequent codes present in the dataset. Similarly for MIMIC-II, LAAT performs better compared to MultiResCNN and improves macro-AUC, micro-AUC, F1\textsubscript{macro}, F1\textsubscript{micro} and P@8 by 1.8\%, 0.5\%, 0.7\%, 2.2\% and 0.6\% respectively. On the other hand JointLAAT does better on infrequent codes than LAAT with the improvement of 0.9\% on the F1\textsubscript{micro}. 

Similarly, \citet{Song2020} propose an Adversarial Generative Model with Hierarchical Tree structure (AGM-HT) for generalised zero-shot learning. AGM-HT exploits the hierarchical structure of ICD codes to generate semantically meaningful features without any labeled data. The study compares their method with the existing state-of-the-art approaches on zero-shot ICD coding \cite{rios-kavuluru-2018-shot}, meta-embedding for long-tailed problem \cite{Liu2019CVPR}, WGAN-GP with classification loss \cite{Xian2018-zeroshot} and with cycle-consistent loss \cite{Felix2018}. Their model improves F1-score from nearly 0 to 20.91\% for the zero-shot codes and increases the AUC score by 3\%. A study by \citet{Donglin2020} demonstrates that disease inference based on symptom representation vectors outperforms other methods based on full texts. The study uses only first three characters of each code, rather than a complete code, and performs multi-label classification for the 50 most common and 100 most common diseases only. Furthermore, the study compares the effect of BiLSTMs with different symptom representations, and finds that the combination of two symptom representations improves the performance of disease inference and achieves AUC of 0.895 and F1 of 0.572 for 50 most common diseases.

\citet{Shaoxiong2020-dilated} proposed DCAN, integrating dilated convolutions, residual connections, and label attention for ICD code assignment. The model gave better performance than the state-of-the-art methods\cite{Prakash2017, Shi2017, Mullenbach2018, Wang2018-joint-embedding, LiFei2020} on top 50 ICD codes prediction by achieving macro F1 of 61.5\% and micro F1 of 67.1\%. \citet{Du2019} proposed  ML-Net deep learning framework which combines a label prediction network with an automated label count prediction mechanism to provide an optimal set of labels. The study experimented on 3 independent corpora in 2 text genres- biomedical literature and clinical notes (used MIMIC-II data). For the diagnosis code assignment task, the ML-CNN-threshold achieved the highest F-score of 0.428 among all models. However, the study has certain limitations as the authors di not perform a thorough hyper-parameter tuning, the label count prediction network does not work well for the labels with a hierarchical structure. Also, the document encoding network that maps the text to a high-dimensional representation can be further improved by using an advanced language representation models. 

\citet{Teng2020} proposed a new end-to-end method called G\textunderscore Coder which consists of Multi-CNN, graph presentation, attention matching and adversarial learning for automatic ICD code assignment. The model outperformed other baseline models, including C-LSTM-Att~\cite{Shi2017}, CAML and DR-CAML~\cite{Mullenbach2018}, and MultiResCNN~\cite{LiFei2020}, by achieving  micro-AUC 0f 0.692 and micro-F1 of 0.933 on top 50 common ICD-9 codes. However, the model does not perform well on infrequent codes. \citet{Moons2020} performed a comparison of deep learning methods (CNN, BiGRU, DR-CAML, MVC-LDA, MVC-RLDA) on two publicly available datasets- MIMIC-III and CodiEsp that represent ICD-9 and ICD-10 coding respectively. Similarly, \citet{Chung2020} used deep learning methods such as CNN, LSTM, GRU and HAN, to tackle the multi-label classification problem in ICD coding. Among all the methods, CNN outperformed with micro F1-measure of 76\% in label-to-chapter, top-50 and top-100 common labels.

\subsection{Comparison of studies}

This section highlights the comparison of selected studies that consider automated ICD coding of discharge summaries. It is difficult to compare different models and find out the best one as each model relies on different parameters and conditions such as type of dataset, data distribution, pre-processing pipeline and evaluation metrics. Therefore, we have listed the above mentioned ML and DL studies in the tabular form and compared against each other if they have used the same dataset, data distribution (train-test split) size and number of codes. The use of algorithms and type of evaluation metric (Micro-average, Macro-average or Weighted measures) are the researchers' choice to use them in their studies as there are no strict measures to be used in a particular study. Therefore, we have not considered the evaluation metric as an important parameter for the comparison of studies. It can be observed in Tables~\ref{tab:comparisonOfStudies1}\nobreakdash--\ref{tab:comparisonOfStudies6} that we have grouped the studies and compared with each other checking if they have used the same dataset, train-test spilt size and number of codes, rather than evaluation metrics. 

Among all the selected studies, the majority of them have experimented with MIMIC-II and MIMIC-III datasets as they are publicly available. Out of 38 selected studies,  3 studies \cite{Perotte2013, Marafino2014, Du2019} have used MIMIC-II dataset only, 16 studies \cite{Ayyar2016, Prakash2017, Berndorfer2017, Catling2018, Samonte2018, Xie2018, Xie2019, Falis2019, Huang2019, Henning2019, Xu-Keyang2019, Donglin2020, Song2020, Teng2020, Shaoxiong2020-dilated, Chung2020} have used MIMIC-III data only, 7 studies \cite{Baumel2018, Mullenbach2018, rios-kavuluru-2018-shot, Min2019, Cao2020, Thanh2020, LiFei2020} have used both MIMIC-II and MIMIC-III dataset, 4 studies \cite{Zeng2019, Aaron2020, Mascio2020-comparative, Moons2020} have used MIMIC-III along with other dataset, 6 studies \cite{subotin2014, Kavuluru2015, Chin2017, Amoia2018, kaur2018, Rios2019, Rajvir2019, Zachariah2020-bert-xml}  used private dataset. Also, a few studies \cite{Prakash2017, Mullenbach2018, Xie2019, Donglin2020, Thanh2020, Cao2020, LiFei2020, Moons2020} have divided MIMIC-III dataset into two common settings, MIMIC-III full and MIMIC-III 50 (or MIMIC-III 100), where MIMIC-III full contains a complete set of ICD codes for 52,722 discharge summaries, MIMIC-III 50 contains the top 50 most frequent codes, and MIMIC-III 100 contains 100 most frequent codes for discharge summaries. However, there might be a slight variation in the number of codes when MIMIC-III full data ares used in the studies, but the process of assigning the codes is same. 

 %In Table\ref{tab:comparisonOfStudies}, the comparison of studies is done individually based on the type of data, train-test split used for experiments, year of publication, algorithms used and performance evaluated using different evaluation metrics. 
 
 Table~\ref{tab:comparisonOfStudies1} gives a comparison of studies that used MIMIC-II dataset. Six studies \cite{Perotte2013, Mullenbach2018, Min2019, Cao2020, Thanh2020, LiFei2020} have used the same train-test split size and number of codes. Among all these six studies, \citet{Thanh2020} give more prominent results using LAAT and JointLAAT model. \citet{Baumel2018} also used the same train-test split size, but with more number of codes; they have also used additional data (MIMIC-III) for training. Similarly, \citet{Du2019} have divided MIMIC-II data into training, validation and testing and have more than 7000 unique codes. Therefore, we have not compared \citet{Baumel2018} and \citet{Du2019} with the above mentioned six studies. The other two studies, \cite{Marafino2014} and \cite{rios-kavuluru-2018-shot}, are individual studies and they are focused towards a dedicated task.
 
 Similarly, Tables~\ref{tab:comparisonOfStudies2}\nobreakdash--\ref{tab:comparisonOfStudies3} show the comparison of studies using MIMIC-III dataset discharge summaries.  Six studies, \cite{Falis2019, Xie2019, Cao2020, Thanh2020, LiFei2020, Moons2020}, have used same data distribution and unique codes. Among them, \citet{Thanh2020} give more prominent results using LAAT and JointLAAT model in all the evaluation metrics except in micro-averaged AUC. Moreover, \citet{Mullenbach2018} have also used the same data split size, but they have higher number of unique codes; therefore, we consider it as a separate study. However, except \citet{Moons2020}, the remaining five studies\cite{Falis2019, Xie2019, Cao2020, Thanh2020, LiFei2020} have compared their models' performance with CAML and DR-CAML method which is developed by \citet{Mullenbach2018}. Therefore, we have found that the majority of studies using MIMIC-III data discharge summaries compare their proposed methods with three state-of-the-art models: Flat \& Hierarchy-based SVM \cite{Perotte2013}, CAML \& DR-CAML \cite{Mullenbach2018}, and HA-GRU \cite{Baumel2018}. 
 
 Table~\ref{tab:comparisonOfStudies4} shows the comparison of studies that experimented either with top 100 or 50 most frequent codes in the study. For top 50 most frequent ICD codes, LAAT and JointLAAT proposed by \cite{Thanh2020} again outperforms among all other models. Similarly, Table~\ref{tab:comparisonOfStudies5} shows the list of studies that experimented with less than 50 codes or that have used other datasets along with MIMIC-III. Lastly, Table~\ref{tab:comparisonOfStudies6} shows the list of studies that have used private datasets and are considered as individual studies.

%$$$$$$$$$$$$$$$$$$$$$$$$

%################## TABLE 1 #######################

\begin{landscape}
\begin{table}
 \caption{Comparison of studies using MIMIC-II dataset, different algorithms and evaluation metrics}
\label{tab:comparisonOfStudies1}

\centering
\definecolor{light-gray}{gray}{0.9}
\begin{tabular}{|p{4cm}|p{0.6cm}|p{3.2cm}|p{3.5cm}|p{11cm}|} 
  %\toprule
  \hline
 \textbf{Train-Test Split}  & \textbf{Year} & \textbf{Study} & \textbf{Algorithms} & \textbf{Performance}   \\ \hline

   %\midrule 
    \multirow{14}{4cm}{D\textsubscript{Total}: 22,815, D\textsubscript{Train}: 20,533, D\textsubscript{Test}: 2,282, Codes: 5031 } & 2013 & \citet{Perotte2013} & Flat SVM & P: 86.7\%; R: 16\%; F-measure: 27.6\% \\
    & & & Hierarchy-based SVM & P: 57.7\%, R: 30.0\%, F-measure: 39.5\% \vspace{1.5mm} \\ \cdashline {2-5}
   
   & 2018 & \citet{Mullenbach2018} & CAML & AUC\textsubscript{mi}: 0.966; AUC\textsubscript{ma}: 0.820;  F1\textsubscript{ma}: 0.048;  F1\textsubscript{mi}: 0.442; P@8:0.523 \\
   & & &  DR-CAML & AUC\textsubscript{mi}: 0.966; AUC\textsubscript{ma}: 0.826;  F1\textsubscript{ma}: 0.049;  F1\textsubscript{mi}: 0.457; P@8: 0.515 \vspace{1.5mm}\\  \cdashline {2-5}

   & 2019 & \citet{Min2019} & DeepLabeler (CNN+D2V) \vspace{1.5mm} & P\textsubscript{mi}: 0.475; R\textsubscript{mi}: 0.258; F-meas\textsubscript{mi}: 0.335 \\  \cdashline {2-5} 
   
   & 2020 & \citet{Cao2020} & HyperCore & AUC\textsubscript{mi}: 0.971; AUC\textsubscript{ma}: 0.885;  F1\textsubscript{ma}: 0.070;  F1\textsubscript{mi}: 0.477; P@8:0.537  \vspace{1.5mm} \\ \cdashline {2-5} 
   
    &  2020 & \citet{Thanh2020} & LAAT \newline JointLAAT & AUC\textsubscript{mi}: 0.973; AUC\textsubscript{ma}: 0.868;   F1\textsubscript{ma}: 0.059;  F1\textsubscript{mi}: 0.486; P@5: 0.649; P@8: 0.550; P@15: 0.397 \newline AUC\textsubscript{mi}: 0.972; AUC\textsubscript{ma}: 0.871;   F1\textsubscript{ma}: 0.068;  F1\textsubscript{mi}: 0.491; P@5: 0.652; P@8: 0.551; P@15: 0.396 \vspace{1.5mm} \\  \cdashline {2-5}

   & 2020 & \citet{LiFei2020} & MultiResCNN & AUC\textsubscript{mi}:$0.968\pm0.001$;  AUC\textsubscript{ma}:$0.850\pm0.002$;  F1\textsubscript{ma}:$0.052\pm0.002$; F1\textsubscript{mi}:$0.464\pm0.002$; P@8: $0.544\pm0.007$ \vspace{1.5mm} \\ \hline
  
     \multirow{3}{4cm}{D\textsubscript{Total}: 4191 Neonatal ICU, D\textsubscript{Total}: 2198 Adult ICU} \vspace{0.3cm} & 2014 & \citet{Marafino2014} & SVM & P: 0.982; R: 0.952; F1: 0.954; Acc: 0.982 (Ventilation classification task)  The study focused on only two diagnoses and two procedures. \vspace{1.5mm}\\ \hline
   
    \multirow{3}{3.8cm}{D\textsubscript{Total}: 22,815, D\textsubscript{Train}: 20,533, D\textsubscript{Test}: 2,282, Codes: 6,527} & 2018 & \citet{Baumel2018} & SVM & F\textsubscript{mi}: 28.13\%; F\textsubscript{mi}: 32.50\% (Rolled- up codes)\\
    & & &CBOW & F\textsubscript{mi}: 30.60\%; F\textsubscript{mi}: 42.06\% (Rolled- up codes)\\
    & & & CNN & F\textsubscript{mi}: 33.25\%; F\textsubscript{mi}: 46.40\% (Rolled- up codes) \\
    & & & HA-GRU & F\textsubscript{mi}: 36.60\%; F\textsubscript{mi}: 53.86\% (Rolled- up codes) \vspace{1.5mm}\\ \hline

     \multirow{3}{4cm}{ D\textsubscript{Train}: 18,822, D\textsubscript{Test}: 1,711, Codes: 3,118 (Group S: all codes occur > 5 times); 3,459 (Group F: codes between 1 to 5 times); 355 (Group Z: never occur)} & 2018 & \citet{rios-kavuluru-2018-shot} & ZACNN & Group S= R@5: 0.135; R@10: 0.247 \\ 
     & & & ZAGCNN &  Group F= R@5: 0.130; R@10: 0.185 \\
     & & &  &  Group Z= R@5: 0.269; R@10: 0.362 \\
     & & & & Harmonic Average= R@5: 0.160; R@10: 0.246 \\
     & & & & The above given results are of ZAGCNN as it gives better performance than ZACNN. \vspace{1.5mm} \\ \hline

    \multirow{3}{3.8cm}{ D\textsubscript{Total}: 22,815, D\textsubscript{Train}: 70\%, D\textsubscript{Valid}: 10\%, D\textsubscript{Test}: 20\%, Codes: 7024} & 2019 & \citet{Du2019} & ML-Net, ML-CNN,  & P: 0.501; R:0.373; F-score:0.428 \\
    & & & ML-HAN & ML-CNN-threshold achieved highest F-score among all models \vspace{3.2mm}\\ \hline

 %\bottomrule
\end{tabular} 

\footnotesize{D\textsubscript{Total}: Total no. of records, D\textsubscript{Train}: Train data, D\textsubscript{Valid}: Validation data, D\textsubscript{Test}:Test data,  P\textsubscript{mi}: Micro-averaged Precision,  R\textsubscript{mi}: Micro-averaged Recall, F1\textsubscript{mi} or F-meas\textsubscript{mi}: Micro-averaged F-measure, AUC\textsubscript{mi}: Micro-averaged F1 score of Area under the ROC curve (AUC), P\textsubscript{ma}: Macro-averaged Precision,  R\textsubscript{ma}: Macro-averaged Recall, F1\textsubscript{ma}: Micro-averaged F1-score, AUC\textsubscript{ma}: Macro-averaged F1 score of Area under the ROC curve (AUC), F1\textsubscript{diag}: Micro- F1 on diagnosis codes,  F1\textsubscript{proc}: Micro-F1 on procedure codes, P:Precision, R: Recall, F1: F1-score, ACC: Accuracy, P@n: Precision@n; R@n: Recall@n.}
\end{table}
\end{landscape}

%################## TABLE 2 #######################

%##########
\begin{landscape}
\begin{table}
\caption{Comparison of studies using MIMIC-III dataset, different  algorithms and evaluation metrics}
\label{tab:comparisonOfStudies2}
\centering
\definecolor{light-gray}{gray}{0.9}

\begin{tabular}{|p{4cm}|p{0.6cm}|p{3.2cm}|p{3.5cm}|p{11cm}|} \hline
  %\toprule
  \textbf{Train-Test Split}  & \textbf{Year} & \textbf{Study} & \textbf{Algorithms} & \textbf{Performance} \\ \hline
   %\midrule 

   \multirow{2}{4cm}{D\textsubscript{Total}: 59,531, Codes: 1,301} & \multirow{2}{0.6cm}{2017} & \multirow{2}{2.2cm}{\citet{Berndorfer2017}} & Flat SVM & P: 55.10\%; R: 33.74\%; F1-score: 39.16\% \\
  & & &  Hierarchy-based SVM & P: 40.08\%; Recall: 41.69\%; F1-score: 39.25\% \vspace{1.5mm} \\  \hline
   
   \multirow{2}{4cm}{D\textsubscript{Train}: 37,016, D\textsubscript{Test}: 1,356, Codes: 4,403 (Group S: all codes occur > 5 times); 4,403 (Group F: codes between 1 to 5 times); 178 (Group Z: never occur) } & 2018 &\citet{rios-kavuluru-2018-shot}  & ZACNN & Group S= R@5: 0.283; R@10: 0.445\\
   & & & ZAGCNN &  Group F= R@5: 0.166; R@10: 0.216 \\
   & & & & Group Z= R@5: 0.428; R@10: 0.495 \\
   & & & & Harmonic Average= R@5: 0.252; R@10: 0.337 \vspace{0.9cm}  \\    \hline

   \multirow{2}{4cm}{D\textsubscript{Total}: 55,172, D\textsubscript{Train}: 38,588, D\textsubscript{Valid}: 5,536, D\textsubscript{Test}: 11,048} & 2018 & \citet{Catling2018} & GRU & P: 0.692; R: 0.705 ; F1: 0.691 \\
   & & & & Results for chapter (level 1) label prediction. RNN text representation improved Weighted F1 for prediction of 19 disease-category labels 0.682-0.701 from 0.662-0.682 using tf-idf. \vspace{1.5mm} \\     \hline

    \multirow{2}{4cm}{D\textsubscript{Train}: 36,998, D\textsubscript{Valid}: 1,632, D\textsubscript{Test}: 3,372, Codes: 8,921} & 2018 & \citet{Mullenbach2018} & CAML & AUC\textsubscript{mi}: 0.986; AUC\textsubscript{ma}: 0.895;  F1\textsubscript{ma}: 0.088;  F1\textsubscript{mi}: 0.539; F1\textsubscript{diag}: 0.524;  F1\textsubscript{proc}: 0.609; P@8: 0.709; P@15: 0.561 \\
    & & & DR-CAML & AUC\textsubscript{mi}: 0.985; AUC\textsubscript{ma}: 0.879;  F1\textsubscript{ma}: 0.086;  F1\textsubscript{mi}: 0.529;  F1\textsubscript{diag}: 0.515;  F1\textsubscript{proc}: 0.595; P@8: 0.690; P@15: 0.548  \vspace{1.5mm} \\  \hline
   
   \multirow{2}{4.5cm}{D\textsubscript{Total}: 49,857, Codes:4,847} & 2018 & \citet{Baumel2018} & SVM & F\textsubscript{mi}: 22.25\%; F\textsubscript{mi}: 53.02\% (Rolled- up codes) \\
   & & & CBOW & F\textsubscript{mi}: 30.02\%; F\textsubscript{mi}: 43.30\% (Rolled- up codes) \\
   & & &  CNN & F\textsubscript{mi}: 40.72\%; F\textsubscript{mi}: 52.64\% (Rolled- up codes) \\
   & & & HA-GRU & F\textsubscript{mi}: 40.52\%; F\textsubscript{mi}: 55.86\% (Rolled- up codes) \vspace{1.5mm} \\  \hline
   
    \multirow{2}{4cm}{D\textsubscript{Total}: 58,976, D\textsubscript{Train}:40k, D\textsubscript{Valid}:7k, D\textsubscript{Test}:12k, Codes:6,984} &  2018 & \citet{Xie2018} &  Tree of sequences LSTM \vspace{0.8cm} & Sensitivity: 0.29; Specificity: 0.33  \\  \hline

    \multirow{2}{4cm}{D\textsubscript{Total}: 59,652, Codes: 6,918, 10-fold cross validation} & 2019 & \citet{Henning2019}& FastText & P:0.58; R: 0.668; F1-score: 0.622 (eFastText-UMLS\textsubscript{cardinality}) \vspace{1.2mm}	  \\    \hline
    
    \multirow{2}{4cm}{D\textsubscript{Total}: 52,962, D\textsubscript{Train}: 47,665, D\textsubscript{Test}: 5,297, Codes: 6,984} & 2019 &  \citet{Min2019} & DeepLabeler (CNN+D2V) \vspace{0.7cm} & P\textsubscript{mi}: 0.486; R\textsubscript{mi}: 0.351; F-meas\textsubscript{mi}: 0.408  \\ \hline
                                    
\hline
   
\end{tabular}

\footnotesize{D\textsubscript{Total}: Total no. of records, D\textsubscript{Train}: Train data, D\textsubscript{Valid}: Validation data, D\textsubscript{Test}:Test data,  P\textsubscript{mi}: Micro-averaged Precision,  R\textsubscript{mi}: Micro-averaged Recall, F1\textsubscript{mi} or F-meas\textsubscript{mi}: Micro-averaged F-measure, AUC\textsubscript{mi}: Micro-averaged F1 score of Area under the ROC curve (AUC), P\textsubscript{ma}: Macro-averaged Precision,  R\textsubscript{ma}: Macro-averaged Recall, F1\textsubscript{ma}: Micro-averaged F1-score, AUC\textsubscript{ma}: Macro-averaged F1 score of Area under the ROC curve (AUC), F1\textsubscript{diag}: Micro- F1 on diagnosis codes,  F1\textsubscript{proc}: Micro-F1 on procedure codes, P: Precision, R: Recall, F1: F1-score, ACC: Accuracy.}
\end{table}
\end{landscape}

%################## TABLE 3 #######################

%##########
\begin{landscape}
\begin{table}
\caption{Comparison of studies using MIMIC-III datasets, different  algorithms and evaluation metrics}
\label{tab:comparisonOfStudies3}
\centering
\definecolor{light-gray}{gray}{0.9}

\begin{tabular}{|p{4cm}|p{0.6cm}|p{3.2cm}|p{3.5cm}|p{11cm}|} \hline
  %\toprule
 \textbf{Train-Test Split}  & \textbf{Year} & \textbf{Study} & \textbf{Algorithms} & \textbf{Performance} \\ \hline
   %\midrule 

  \multirow{14}{4cm}{D\textsubscript{Train}: 47,719, D\textsubscript{Valid}: 1,631, \newline D\textsubscript{Test}: 3,372, Codes: 8,929 (Approx.)}  & 2019 & \citet{Falis2019} & Ontological Attention & P\textsubscript{mi}: 0.617; R\textsubscript{mi}: 0.514; F1\textsubscript{mi}: 0.560; P@8: 0.727; P\textsubscript{ma}: 0.192; R\textsubscript{ma}: 0.341; F1\textsubscript{ma}: 0.245; P@8: 0.681 \vspace{1.5mm} \\ \cdashline {2-5}
   
   & 2019 & \citet{Xie2019} & MSATT-KG & AUC\textsubscript{ma}:91.0\%; AUC\textsubscript{mi}: 99.2\%; F1\textsubscript{ma}: 9.0\% ; F1\textsubscript{mi}: 55.3\%; F1\textsubscript{diag}:54.0\%; F1\textsubscript{proc}: 62.3\%; P@8: 72.8\%; P@15:58.1\% \vspace{1.5mm} \\ \cdashline {2-5}
   
    & 2020 & \citet{Cao2020} &   Hypercore & AUC\textsubscript{mi}: 0.989; AUC\textsubscript{ma}: 0.930;  F1\textsubscript{ma}: 0.090;  F1\textsubscript{mi}: 0.551; P@8: 0.722; P@15: 0.579  \vspace{1.5mm}\\ \cdashline {2-5}
   
   &  2020 & \citet{Thanh2020} &  LAAT \newline JointLAAT & AUC\textsubscript{ma}: 0.919; AUC\textsubscript{mi}: 0.988;  F1\textsubscript{ma}: 0.099;  F1\textsubscript{mi}: 0.575; P@5: 0.813; P@8: 0.738; P@15: 0.591 \newline  AUC\textsubscript{ma}: 0.921; AUC\textsubscript{mi}: 0.988;  F1\textsubscript{ma}: 0.107;  F1\textsubscript{mi}: 0.575; P@5: 0.806; P@8: 0.735; P@15: 0.590 \vspace{1.5mm} \\ \cdashline {2-5}

    & 2020 &\citet{LiFei2020}&  MultiResCNN & AUC\textsubscript{ma}:$0.910\pm0.002$, AUC\textsubscript{mi}:$0.986\pm0.001$, F1\textsubscript{ma}:$0.085\pm0.007$, F1\textsubscript{mi}:$0.552\pm0.005$, P@8:$0.734\pm0.002$, P@15: $0.584\pm0.001$ \vspace{1.5mm} \\ \cdashline {2-5}
   
    & 2020 & \citet{Moons2020} & CNN, GRU, DR-CAML, MVC-LDA, MVC-RLDA  &   AUC\textsubscript{mi}: 90.02; F1\textsubscript{mi}: 59.75 (Procedure); F1\textsubscript{mi}:51.60 (Diagnosis); F1\textsubscript{mi}:55.03(Both), P@5:69.77 (MVC-LDA outperforms other models and data includes only discharge summaries) \vspace{2mm} 
   
     F1\textsubscript{mi}: 58.12 (Procedure); AUC\textsubscript{mi}: 89.93; F1\textsubscript{mi}:50.70 (Diagnosis); F1\textsubscript{mi}:51.97(Both),  P@5:68.53 (MVC-LDA outperforms other models and data includes discharge summary and other notes (radiology, nursing notes etc.) \vspace{1.5mm} \\ \hline
    
   \multirow{2}{4cm}{D\textsubscript{Train}: 46,157, D\textsubscript{Valid}: 3,280, D\textsubscript{Test}: 3,285, Codes: 6,916} & 2020 & \citet{Song2020} & ZAGRNN & P\textsubscript{mi}: 0.5806;  R\textsubscript{mi}: 0.4494; F1\textsubscript{mi}: 0.5066; AUC\textsubscript{mi}:0.9667; P\textsubscript{ma}: 0.3091;  R\textsubscript{ma}: 0.2557; F1\textsubscript{ma}: 0.2799; AUC\textsubscript{ma}: 0.9403 \\
   
   & & & ZAGRNN + L\textsubscript{LDAM} & P\textsubscript{mi}: 0.5606;  R\textsubscript{mi}: 0.4714; F1\textsubscript{mi}: 0.5122; AUC\textsubscript{mi}:0.9670; P\textsubscript{ma}: 0.3172;  R\textsubscript{ma}: 0.2806; F1\textsubscript{ma}: 0.2978; AUC\textsubscript{ma}: 0.9408 \\
   & &  & & The above mentioned results are for seen codes. However, for zero-shot codes F1\textsubscript{mi} improved from 0 to 20.91\% and AUC\textsubscript{mi} improved by 3\% and reached 92.18\% \vspace{1.5mm} \\  \hline
   
  \multirow{2}{4cm}{D\textsubscript{Total}: 59,542  D\textsubscript{Train}: 70\%, D\textsubscript{Valid}: 10\%, D\textsubscript{Test}: 20\% (Label-to-Chapter)}  & 2020 & \citet{Chung2020} & CNN, LSTM, GRU, HAN & F1\textsubscript{mi}:76\% (CNN outperforms other models) \vspace{0.9cm} \\ \hline

\end{tabular}

\footnotesize{D\textsubscript{Total}: Total no. of records, D\textsubscript{Train}: Train data, D\textsubscript{Valid}: Validation data, D\textsubscript{Test}:Test data,  P\textsubscript{mi}: Micro-averaged Precision,  R\textsubscript{mi}: Micro-averaged Recall, F1\textsubscript{mi} or F-meas\textsubscript{mi}: Micro-averaged F-measure, AUC\textsubscript{mi}: Micro-averaged F1 score of Area under the ROC curve (AUC), P\textsubscript{ma}: Macro-averaged Precision,  R\textsubscript{ma}: Macro-averaged Recall, F1\textsubscript{ma}: Micro-averaged F1-score, AUC\textsubscript{ma}: Macro-averaged F1 score of Area under the ROC curve (AUC), F1\textsubscript{diag}: Micro- F1 on diagnosis codes,  F1\textsubscript{proc}: Micro-F1 on procedure codes, P: Precision, R: Recall, F1: F1-score, ACC: Accuracy, HL: Hamming Loss, JS: Jaccard Similarity}
\end{table}
\end{landscape}

%####################TABLE 4######################################

\begin{landscape}
\begin{table}
\caption{Comparison of studies using MIMIC-III dataset with top (n=50, 100) codes, different algorithms and evaluation metrics}
\label{tab:comparisonOfStudies4}
\centering
\definecolor{light-gray}{gray}{0.9}
\begin{tabular}{|p{1.8cm}|p{3.8cm}|p{0.6cm}|p{2.4cm}|p{3.2cm}|p{10.2cm}|} \hline
  %\toprule
 \textbf{Dataset} & \textbf{Train-Test Split}  & \textbf{Year} & \textbf{Study} & \textbf{Algorithms} & \textbf{Performance}   \\  \hline

  % \midrule 

  \multirow{7}{2cm}{\bfseries MIMIC-III using top 100 codes } &  \multirow{2}{4cm}{D\textsubscript{Train}: 80\%, D\textsubscript{Valid}: 10\%, D\textsubscript{Test}: 10\%} & 2017 & \citet{Prakash2017} & A-MemNN & AUC\textsubscript{ma}: 0.720; Average P@5: 0.29; HL:0.11 \\
  & & & & C-MemNN &  AUC\textsubscript{ma}: 0.767; Average P@5: 0.32; HL:0.05 \vspace{1.5mm} \\ \cline{2-6}
  
  & \multirow{2}{3.8cm}{D\textsubscript{Total}: 46,715} & 2020 & \citet{Donglin2020} & BiLSTMs & P\textsubscript{mi}:0.496; R\textsubscript{mi}:0.564; F1\textsubscript{mi}:0.528; AUC\textsubscript{mi}:0.87; P\textsubscript{ma}:0.464; R\textsubscript{ma}:0.463; F1\textsubscript{ma}:0.448; AUC\textsubscript{ma}:0.818 \\
  & & &  & & The results mentioned are of BiLSTMs + SymVec (TF-IDF+Word2Vec) \vspace{1mm} \\ \cline{2-6}
  
  & \multirow{2}{3.7cm}{D\textsubscript{Total}: 59,542  D\textsubscript{Train}: 70\% D\textsubscript{Valid}: 10\%, D\textsubscript{Test}: 20\%} & 2020 & \citet{Chung2020} & CNN, LSTM, GRU, HAN  &  F1\textsubscript{mi}:51.4\% (3-digit); F1\textsubscript{mi}:50.2\%\% (4-digit) F1\textsubscript{mi}:55.2\% (5-digit) CNN outperforms other models \vspace{1.5mm}  \\

   %\specialrule{.2em}{.1em}{.1em}
  
   \multirow{16}{2cm}{\bfseries MIMIC-III using top 50 codes } & \multirow{2}{3.5cm}{D\textsubscript{Train}: 80\%, D\textsubscript{Valid}: 10\%, D\textsubscript{Test}: 10\%} & 2017 & \citet{Prakash2017} & A-MemNN & AUC\textsubscript{ma}: 0.804; Average P@5: 0.40; HL:0.02 \\
   & & & & C-MemNN &  AUC\textsubscript{ma}: 0.833; Average P@5: 0.42; HL:0.01 \vspace{1.5mm}  \\ \cline{2-6}
   
   & \multirow{8}{3.5cm}{D\textsubscript{Train}: 8,067, D\textsubscript{Valid}: 1,574, D\textsubscript{Test}: 1,730} & 2018 & \citet{Mullenbach2018} & CAML, \newline DR-CAML & AUC\textsubscript{mi}: 0.909; AUC\textsubscript{ma}: 0.875;  F1\textsubscript{ma}: 0.532;  F1\textsubscript{mi}: 0.614; P@5:0.609,  \newline AUC\textsubscript{mi}: 0.916; AUC\textsubscript{ma}: 0.884;  F1\textsubscript{ma}: 0.576;  F1\textsubscript{mi}: 0.633; P@5: 0.618  \vspace{1.5mm} \\  \cdashline{3-6}
   
   & & 2019 & \citet{Xie2019} & MSATT-KG & AUC\textsubscript{ma}:91.4\%; AUC\textsubscript{mi}: 93.6\%; F1\textsubscript{ma}: 63.8\% ; F1\textsubscript{mi}: 68.4\%; P@5: 64.4\% \vspace{1.5mm} \\ \cdashline{3-6}
   
   & & 2020 & \citet{Cao2020} & Hypercore & AUC\textsubscript{mi}: 0.929; AUC\textsubscript{ma}: 0.895;  F1\textsubscript{ma}: 0.609;  F1\textsubscript{mi}: 0.663; P@5: 0.632 \vspace{1.5mm} \\ \cdashline{3-6}
   
   & & 2020 &\citet{Thanh2020}  & LAAT & AUC\textsubscript{ma}: 0.925; AUC\textsubscript{mi}: 0.946;  F1\textsubscript{ma}: 0.666;  F1\textsubscript{mi}: 0.715; P@5: 0.675; P@8: 0.547; P@15: 0.357 \\
   & & & & JointLAAT & AUC\textsubscript{ma}: 0.925; AUC\textsubscript{mi}: 0.946;  F1\textsubscript{ma}: 0.661;  F1\textsubscript{mi}: 0.716; P@5: 0.671; P@8: 0.546; P@15: 0.357 \vspace{1.5mm} \\ \cdashline{3-6}

   & & 2020 & \citet{Moons2020} & CNN, GRU, DR-CAML, MVC-LDA, MVC-RLDA \vspace{1mm} &  F1\textsubscript{mi}: 67.86; F1\textsubscript{ma}: 63.74; AUC\textsubscript{mi}: 93.47; P@5:63.48 (DR-CAML outperforms other models)  \\ \cdashline{3-6}
   
   & & 2020 & \citet{LiFei2020} & MultiResCNN & AUC\textsubscript{ma}:$0.899\pm0.004$, AUC\textsubscript{mi}:$0.928\pm0.002$, F1\textsubscript{ma}:$0.606\pm0.0011$, F1\textsubscript{mi}:$0.670\pm0.003$, P@5: $0.641\pm0.001$ \vspace{1.5mm}\\ \cline{2-6}
    
   & \multirow{2}{3.5cm}{D\textsubscript{Total}: 46,364} & 2020 &\citet{Donglin2020} & BiLSTMs & P\textsubscript{mi}:0.519; R\textsubscript{mi}:0.638; F1\textsubscript{mi}:0.572; AUC\textsubscript{mi}:0.859; P\textsubscript{ma}:0.508; R\textsubscript{ma}:0.568; F1\textsubscript{ma}:0.522; AUC\textsubscript{ma}:0.823 \vspace{1.5mm} \\  \cline{2-6}
   
   & \multirow{2}{3.7cm}{D\textsubscript{Total}: 46,552, D\textsubscript{Train}: 43,000 D\textsubscript{Valid}: 1,800, D\textsubscript{Test}: 1,752} & 2020 &\citet{Teng2020} &G\textunderscore Coder & F1\textsubscript{mi}: 0.692; AUC\textsubscript{mi}: 0.933; P@5: 0.653 \vspace{0.8cm} \\  \cline{2-6}
   
 & \multirow{2}{3.7cm}{D\textsubscript{Train}: 8,066, D\textsubscript{Valid}: 1,573, D\textsubscript{Test}: 1,729} & 2020 & \citet{Shaoxiong2020-dilated} & DCAN & AUC\textsubscript{ma}:$90.2\pm0.6$, AUC\textsubscript{mi}:$93.1\pm0.1$, F1\textsubscript{ma}: $61.5\pm0.7$, F1\textsubscript{mi}:$67.1\pm0.1$, P@8: $64.2\pm0.2$ \vspace{1.5mm} \\ \cline{2-6}
 
 &  \multirow{2}{3.7cm}{D\textsubscript{Total}: 59,542  D\textsubscript{Train}: 70\% D\textsubscript{Valid}: 10\%, D\textsubscript{Test}: 20\%} &2020 & \citet{Chung2020} \vspace{2mm}  & CNN, LSTM, GRU, HAN  & F1\textsubscript{mi}:57.5\% (3-digit); F1\textsubscript{mi}:59.5\%\% (4-digit) F1\textsubscript{mi}:67.4\% (5-digit) CNN outperforms other models  \\ 
    
\hline
 
\end{tabular}

\footnotesize{D\textsubscript{Total}: Total no. of records, D\textsubscript{Train}: Train data, D\textsubscript{Valid}: Validation data, D\textsubscript{Test}:Test data,  P\textsubscript{mi}: Micro-averaged Precision,  R\textsubscript{mi}: Micro-averaged Recall, F1\textsubscript{mi} or F-meas\textsubscript{mi}: Micro-averaged F-measure, AUC\textsubscript{mi}: Micro-averaged F1 score of Area under the ROC curve (AUC), P\textsubscript{ma}: Macro-averaged Precision,  R\textsubscript{ma}: Macro-averaged Recall, F1\textsubscript{ma}: Micro-averaged F1-score, AUC\textsubscript{ma}: Macro-averaged F1 score of Area under the ROC curve (AUC), F1\textsubscript{diag}: Micro- F1 on diagnosis codes,  F1\textsubscript{proc}: Micro-F1 on procedure codes, P: Precision, R: Recall, F1: F1-score, ACC: Accuracy.}
\end{table}
\end{landscape}

%################################TABLE 5 ###############################
\begin{landscape}
\begin{table}
\caption{Comparison of studies using MIMIC-III and other datasets, algorithms and evaluation metrics}
\label{tab:comparisonOfStudies5}
\centering
\definecolor{light-gray}{gray}{0.9}
\begin{tabular}{|p{1.5cm}|p{4cm}|p{0.6cm}|p{2cm}|p{3cm}|p{10.5cm}|} \hline
  %\toprule
 \textbf{Dataset} & \textbf{Train-Test Split}  & \textbf{Year} & \textbf{Study} & \textbf{Algorithms} & \textbf{Performance}   \\  \hline

  % \midrule 

  \multirow{14}{1.5cm}{\bfseries MIMIC-III Less than 50 codes } & D\textsubscript{Train}: 39,541, D\textsubscript{Valid}:13,181, Codes: Top 19 & 2016 & \citet{Ayyar2016} \vspace{1.5mm} & LSTM & P:0.799; R:0.685; F1-score:0.708\\ \cline{2-6}
  
 &  D\textsubscript{Total}: 49,857, Codes:19 & 2018 & \citet{Samonte2018}  & EnHANs & P: 0.910; R: 0.540; F1-score: 0.500; Acc: 0.900 \newline The results described in the analysis section are not matching with the result table provided in the study \vspace{1.5mm} \\  \cline{2-6} 
 
  & D\textsubscript{Train}: 31,155, D\textsubscript{Valid}: 4,484, D\textsubscript{Test}: 9,020, Codes: 32 (mapped from ICD-9 to ICD-10 ) & 2019 & \citet{Xu-Keyang2019} & Ensemble-based (Text-TF-IDF-CNN) & F1\textsubscript{mi}: 0.7633; F1\textsubscript{ma}: 0.6867; AUC\textsubscript{mi}: 0.9541; AUC\textsubscript{ma}: 0.9337; JSC: 0.1806 (Text data) and 0.3105 (tabular data) \newline The combined model of Text-TF-IDF-CNN, Label Smoothing (LS), diagnosis-based ranking (DR) and Tabular data (TD) achieves highest score among all the model tested in the study. \vspace{1.5mm} \\  \cline{2-6} 
  
  & \multirow{2}{4cm}{D\textsubscript{Total}: 52,726, Top-10-code: 40,562, Top-50-code: 49,354, Top-10-cat: 44,419, Top-50-cat:51,034} & 2019 & \citet{Huang2019} & CNNs,\newline LSTM RNNs, \newline GRU RNNs & Top-10-code: P:0.7502; R:0.6519; F1:0.6957; ACC:0.8967 (GRU RNNs)  \newline Top-10-cat: P:0.7580; R:0.6941; F1:0.7233; ACC:0.8588 (GRU RNNs) \newline AUC\textsubscript{ma}:0.8599; P@5:0.8109; HL:0.0645 (GRUs) \vspace{0.5cm}\\ \hline

%   \specialrule{.2em}{.1em}{.1em}

%\multirow{10}{1.5cm}{\bfseries MIMIC-III Nursing notes} & \multirow{9}{4cm}{ D\textsubscript{Total}: 223,556} & 2019 & \citet{Gangavarapu2019-TAGS}&	TAGs model using MLP, LSTM and CNN & ACC:$0.8130\pm0.0005$; AUROC:$0.7817\pm0.0023$; AUPRC:$0.6291\pm0.0027$; MCC:$0.5704\pm0.0020$; F1:$0.6803\pm0.0024$; CE:$18.1300\pm0.1088$; LRL:$0.4124\pm0.0047$ \vspace{1.5mm} \\ \cdashline {3-6}
   
%    & & 2020 & \citet{GANGAVARAPU2020} & TAGs model using kNN, MLP, kNN as OvR, LR as OvR, SVM as OvR, RF, HVE, SE \vspace{1.5mm} & ACC:$0.8239\pm0.0011$; AUROC:$0.7868\pm0.0011$; AUPRC:$0.6476\pm0.0011$; MCC:$0.5953\pm0.0018$; F1:$0.6981\pm0.0016$; CE:$18.2849\pm0.0643$; LRL:$0.3978\pm0.0021$\\ \cdashline {3-6}
    
%    & & \rowcolor{light-gray} 2020 & \citet{Jayasimha2020} & EnTAGs approach using CNN, LSTM, Cascaded CNN-LSTM, GRU \vspace{1.5mm} & ACC:$0.8282\pm0.0023$; AUROC:$0.81578\pm0.0019$; AUPRC:$0.7089\pm0.0046$; MCC:$0.6368\pm0.0042$; F1:$0.7562\pm0.0021$; CE:$17.68\pm0.0566$; LRL:$0.3392\pm0.0036$ (EnTAGs with TW-NMF) \\ 
    
%    & & 2020 &\citet{Gangavarapu2020-FarSight} & MLP, ConvNet, LSTM, Bi-LSTM,Conv-LSTM, Seg-GRU \vspace{1.5mm} & ACC:$0.8343\pm0.0031$; MCC:$0.6459\pm0.0073$; F1:$0.7602\pm0.0068$; AUPRC:$0.7170\pm0.0045$;   AUROC: $0.8192\pm0.0046$ (Group code prediction of nursing notes aggregated using FarSight)\\ 
  
%  \specialrule{.2em}{.1em}{.1em}
  
  \multirow{9}{1.5cm}{\bfseries MIMIC-III used with other datasets } & \multirow{2}{4cm}{D\textsubscript{Total}: 58,929, Codes:6,984, D\textsubscript{Train}: 47,665, D\textsubscript{Test}: 5,297  D\textsubscript{Total}:12,208342 (BioASQ3), MeSH labels:27,301} &  2019 &\citet{Zeng2019} & Deep transfer learning   & P\textsubscript{mi}: 0.483; R\textsubscript{mi}: 0.371; F\textsubscript{mi}: 0.42 \vspace{1.2cm} \\  \cline{2-6} 
  
  & \multirow{2}{4cm}{D\textsubscript{Total}:52,691 (MIMIC-III), D\textsubscript{Total}: 193,677 (PHS)} & 2020 & \citet{Aaron2020} &  UNITE & AUC: 0.91 (PHS) and 0.92 (MIMIC) over six diseases, comparable to LR and MLP \\  \cline{2-6} 
  
  & \multirow{2}{4cm}{D\textsubscript{Total}: 53,423 (MIMIC-III), 10-Fold cross validation} & 2020 & \citet{Mascio2020-comparative} & ANN, CNN, RNN BiLSTM & F1-score\textsubscript{averaged}: 94.5\% (Status); F1-score\textsubscript{averaged}: 97.9\% (Temporality); F1-score\textsubscript{averaged}:98.7\% (ShARe Negation);  F1-score\textsubscript{averaged}:97.3\% (ShARe Uncertainity) \newline Custom Bi-LSTM model outperformed all other models using MIMIC dataset and Word2Vec embedding\\ \hline

 % & \multirow{2}{4cm}{D\textsubscript{Total}: 53,423 (MIMIC-III), 10-Fold cross validation} & 2020 & \citet{Mascio2020-comparative} \vspace{0.5cm} & SVM, ANN, CNN, RNN BiLSTM & \\ 

\end{tabular}

\footnotesize{D\textsubscript{Total}: Total no. of records, D\textsubscript{Train}: Train data, D\textsubscript{Valid}: Validation data, D\textsubscript{Test}:Test data,  P\textsubscript{mi}: Micro-averaged Precision,  R\textsubscript{mi}: Micro-averaged Recall, F1\textsubscript{mi} or F-meas\textsubscript{mi}: Micro-averaged F-measure, AUC\textsubscript{mi}: Micro-averaged F1 score of Area under the ROC curve (AUC), P\textsubscript{ma}: Macro-averaged Precision,  R\textsubscript{ma}: Macro-averaged Recall, F1\textsubscript{ma}: Micro-averaged F1-score, AUC\textsubscript{ma}: Macro-averaged F1 score of Area under the ROC curve (AUC), F1\textsubscript{diag}: Micro- F1 on diagnosis codes,  F1\textsubscript{proc}: Micro-F1 on procedure codes, P: Precision, R: Recall, F1: F1-score, ACC: Accuracy}
\end{table}
\end{landscape}

%#################TABLE 6 ########################################

\begin{landscape}
\begin{table}
\caption{Comparison of studies using different datasets, algorithms and evaluation metrics}
\label{tab:comparisonOfStudies6}
\centering
\definecolor{light-gray}{gray}{0.9}
\begin{tabular}{|p{3cm}|p{4cm}|p{0.6cm}|p{2cm}|p{3cm}|p{9.2cm}|} \hline
  %\toprule
 \textbf{Dataset} & \textbf{Train-Test Split}  & \textbf{Year} & \textbf{Study} & \textbf{Algorithms} & \textbf{Performance}   \\  \hline

  % \midrule 
                
 \multirow{6}{3cm}{\bfseries Australian hospital medical records} & \multirow{2}{4cm}{D\textsubscript{Total}: 190 and 235, Train-Test: 80-20\%} & 2018 & \citet{kaur2018}   & SVM, Naive Bayes, Decision Tree, kNN, RF, AdaBoost, MLP & P: 0.9206; R: 0.8505; F-score: 0.8730; Acc: 0.7920; HL: 0.0877; JS: 0.7453 (For data190 using Decision Tree)  \newline P: 0.9239; R: 0.9201; F-score: 0.9141; Acc: 0.8611; HL: 0.0945; JS: 0.8294 (For data235 using AdaBoost) \\ \cline{2-6}

  & \multirow{2}{3cm}{D\textsubscript{Total}: 190 and 380, Codes: 420 } & 2019 & \citet{Rajvir2019}   & Binary Relevance, Label Power-set, ML-kNN & P\textsubscript{ma}: 0.3723; R\textsubscript{ma}: 0.3739; F1\textsubscript{ma}: 0.3717; P\textsubscript{mi}: 0.9547; R\textsubscript{mi}: 0.9476; F1\textsubscript{mi}: 0.9511; HL: 0.0016; JS: 0.9121 \newline Repetitive task gives better result than other two tasks using Binary Relevance classifier\\ \hline
 
 % \specialrule{.2em}{.1em}{.1em}              

  \multirow{4}{3cm}{\bfseries 10 Healthcare providers and covers 17 months of data} & Train/dev: 16 months data, Test: 1 month data, Codes: 3000  & 2018  & \citet{Amoia2018} \vspace{0.7cm} & LR+CNN & P\textsubscript{mi}: 0.681; R\textsubscript{mi}: 0.616; F1\textsubscript{mi}: 0.646 \\ \hline
 
 % \specialrule{.2em}{.1em}{.1em}
  
  \multirow{6}{3cm}{\bfseries University of Kentucky (UKY) medical center} & \multirow{2}{4cm}{D\textsubscript{Total}:71,463 EMRs, Codes: 7,485} & 2019 & \citet{Rios2019}  \vspace{1mm} & Transfer Learning (CNNs)  &  F\textsubscript{mi}: 0.567; F\textsubscript{ma}: 0.286 \vspace{0.5cm} \\ \cline{2-6}
  
 & \multirow{2}{4cm}{D\textsubscript{Total}: 827, D\textsubscript{Train}: 727,  D\textsubscript{Test}:100, Codes:56 (UKSmall)  \newline D\textsubscript{Total}: 71,463, D\textsubscript{Train}: 66,463, D\textsubscript{Valid}:2000, D\textsubscript{Test}: 3000, Codes: 1231(UKLarge) } & 2015 & \citet{Kavuluru2015} \vspace{1.5cm} & Binary relevance, copy transformation, ECC & CMC data: P\textsubscript{mi}: 0.88, R\textsubscript{mi}: 0.82, F\textsubscript{mi}: 0.85 (ECC + SVM), \newline UKSmall:  F\textsubscript{mi}: 0.44, F\textsubscript{ma}:0.32  (BNS+OTS+MLPTO), \newline UKLarge: F\textsubscript{mi}: 0.479, F\textsubscript{ma}: 0.211 (LR+L2R+NERC)  \\ \hline

 % \specialrule{.2em}{.1em}{.1em}
  
   \multirow{4}{3cm}{\bfseries CodiEsp (English version)} & \multirow{2}{4cm}{D\textsubscript{Total}:1000,  D\textsubscript{Train}: 500, D\textsubscript{Valid}: 250, D\textsubscript{Test}: 250, Codes: 1,767} & 2020 & \citet{Moons2020} & CNN, GRU, DR-CAML, MVC-LDA, MVC-RLDA  & F1\textsubscript{mi}: 12.52 (CNN); F1\textsubscript{ma}: 11.03(GRU); AUC\textsubscript{mi}: 50.54(GRU); P@5:7.96(CNN) Most categories led to lower performance of all models due to insufficient amount of training data.  \\ \hline
  
  \multirow{4}{3cm}{\bfseries The Tri-Service General Hospital, Taipei, Taiwan} & \multirow{2}{4cm}{D\textsubscript{Total}: 103,390, 5-fold cross validation} & 2017 & \citet{Chin2017}   \vspace{1cm} & CNN & F-measure: 0.9086; AUC:0.9696  \\ \hline

%\specialrule{.2em}{.1em}{.1em}
  
  \multirow{4}{3cm}{\bfseries The NYU Langone Hospital EHR} & \multirow{2}{4cm}{D\textsubscript{Total}:7.5 million notes,  D\textsubscript{Train}: 70\%, D\textsubscript{Valid}: 10\%, D\textsubscript{Test}: 20\%, Codes: 2,292} & 2020 & \citet{Zachariah2020-bert-xml} & BERT-XML  & AUC\textsubscript{mi}:0.970; AUC\textsubscript{ma}: 0.927 For EHR BERT Big model with maximum input length of 1024. \vspace{5mm} \\ \hline
  
   %\specialrule{.2em}{.1em}{.1em}

 \multirow{4}{3cm}{\bfseries Private data (Individual clinical records)} & D\textsubscript{Total}:28,536, Thousand EHRs for development testing and evaluation, each, and rest for training along with 175,798 outpatient surgery EHRs with ICD-9 procedures codes; Codes:5,650 (Unique PCS codes) & 2014  & \citet{subotin2014} \vspace{0.7cm}& Two-level hierarchical classification & MRR: 0.572 (All data, features and l\textsubscript{2} regularization \\ \hline

\end{tabular}

\footnotesize{D\textsubscript{Total}: Total no. of records, D\textsubscript{Train}: Train data, D\textsubscript{Valid}: Validation data, D\textsubscript{Test}:Test data,  P\textsubscript{mi}: Micro-averaged Precision,  R\textsubscript{mi}: Micro-averaged Recall, F1\textsubscript{mi} or F-meas\textsubscript{mi}: Micro-averaged F-measure, AUC\textsubscript{mi}: Micro-averaged F1 score of Area under the ROC curve (AUC), P\textsubscript{ma}: Macro-averaged Precision,  R\textsubscript{ma}: Macro-averaged Recall, F1\textsubscript{ma}: Micro-averaged F1-score, AUC\textsubscript{ma}: Macro-averaged F1 score of Area under the ROC curve (AUC), F1\textsubscript{diag}: Micro- F1 on diagnosis codes,  F1\textsubscript{proc}: Micro-F1 on procedure codes, P:Precision, R: Recall, F1: F1-score, ACC: Accuracy, HL: Hamming Loss, AUROC: Area under the ROC curve, JS: Jaccard Similarity, ECC: Ensemble of classifier chains}
\end{table}
\end{landscape}
%$$$$$$$$$$$$$$$$$$$$$***************************$$$$$$$$$$$$$$$$$$$$$$$
%$$$$$$$$$$$$$$$$$$$$$***************************$$$$$$$$$$$$$$$$$$$$$$$

\section{Future research directions}

Several research gaps and limitations found in the studies are addressed in this literature review. This section highlights various future research directions where a considerable efforts are required to develop an automated ICD coding system. 

\begin{enumerate}

    \item Data source: While conducting this review, we have observed that one of the main challenges in developing an automated ICD coding system is the lack of publicly available benchmark \textit{Gold Standard} dataset. The \textit{Gold Standard} dataset is created by human experts who have a good knowledge of medical terminologies, clinical classification systems, and coding rules and guidelines. To the best of our knowledge, there are a few datasets freely available, such as i2b2\footnote{\url{https://www.i2b2.org/NLP/DataSets/}}, and  PhysioNet\footnote{\url{https://physionet.org/about/database/}} that includes clinical reports such as discharge summaries, nursing notes, progress notes, radiology reports, and pathology reports. However, the data annotated in PhysioNet (MIMIC-II and MIMIC-III) are based on the US classification system ICD-9-CM which is no longer used since the adoption of ICD's 10\textsuperscript{th} revision. Yet, till date, the majority of the studies are predicting ICD-9-CM codes due to the limited data resources. Out of 37 selected studies, only 1 study \cite{Xu-Keyang2019} has mapped 32 ICD-9 codes extracted from MIMIC-III data to ICD-10 codes.  Researchers are just aiming to improve the performance of their model using ICD-9-CM annotated dataset without considering the change in classification version. ICD-9 codes are quite different from ICD-10 codes insense that there is a higher number of diagnoses and procedure codes in ICD-10 than in ICD-9-CM, ICD-10 has alphanumeric codes instead of numeric, the order of some chapters is changed, some titles have been renamed, and conditions have been grouped differently. After being aware of all these changes, the researchers are still developing automated ICD coding system based on ICD-9-CM codes. Therefore, this is one of the main problems that needs to be resolved by creating a benchmark dataset annotated with the latest version of the classification system that can be used for future research purposes.
    
    \item Crowd-sourcing platform: Another possible research area is in developing a crowd-sourcing clinical coding and classification platform where the experts can guide and share their views, ideas and knowledge with the less experienced coders and researchers. A study by \citet{Searle2020} has found that frequently assigned codes in  MIMIC-III data display signs of undercoding up to 35\%. No other study has attempted to validate the MIMIC-III data due to time consuming factor and costly nature of the endeavour. For example, if two clinical coders, worked 38 hours a week re-coding all 52,726 admission notes at a rate of 5 minutes and \$3 per document, that would amount to approximately \$316,000(US) and approximately 115 weeks to create a gold standard dataset \cite{Searle2020}. Therefore, if a crowdsourced  knowledge-based platform was created, then the problem of overcoding, undercoding and lack of data sources using the latest coding version could be resolved. Also, this would help in developing an automated ICD coding system in real-time hospital settings. 
    
     \item National adoption and coding rules:  Many countries follow their own specific classification system such as Australia, Canada and the US. There are a few codes that are coded in a specific country only. For example, there are more types of spiders in Australia than in Canada and the US. Therefore, research studies conducted in different countries may have another new challenge if the data are coded in different classification version. Apart from that, various studies have focused on automated ICD code prediction despite considering the use of coding rules.

    \item Reducing the complex problem: In this review, we have also found that a few studies have either truncated the codes to n-digits (3 or 4) only, predicted only top 50 or 100 most frequent codes \cite{Prakash2017, Donglin2020} or removed the rare occurring codes from the data in order to reduce complexity of the problem. One of the reasons behind this is lower number of reports or the rare codes in the reports. Thus, researchers should consider using multi-modal data or reports in order to reduce the complexity. 
    
    \item Transfer learning approach for automated ICD coding: In many machine learning methods, the training and testing data are drawn from the same feature space with the same distribution. However, in the real-world applications, it is difficult to collect sufficient training data to train a model. In such cases, transfer learning can learn from one related task and apply that knowledge to a target task. This approach has been proven very effectively and applied widely in biomedical research. A few studies, \cite{Zeng2019} and \cite{Rios2019}, have applied transfer learning for automated ICD-9 coding and improved classification performance. Thus, researchers may investigate various transfer learning approaches for automated ICD coding task.

    \item Active learning and reinforcement learning approaches for clinical classification problems: One of the major challenges in using machine learning or deep learning approaches is to train the model with the lower number of reports, imbalanced class or rare class. Active learning  has helped to discover rare class~\cite{Rareclass2013}, imbalanced class and other biomedical classification problems \cite{classification-activeLearning}. Similarly, reinforcement learning has also proven to be suitable for imbalanced data classification problem  as it can pay more attention to minority classes by giving higher rewards to them \cite{reinforce-ImbalanceProblem}. Active learning and reinforcement learning can also be beneficial where unlabeled data can be obtained easily as the annotation of clinical reports is difficult, laborious and expensive \cite{Mujtaba2019}. Therefore, the researchers could also investigate various active learning and reinforcement learning algorithms for clinical report classification problems and improve the efficiency and classification model performance. 
    
\end{enumerate}

\section{Conclusions}
 
This study provides a comprehensive overview of automated ICD coding system based on discharge summaries. In this systematic literature review, we have addressed six research questions. A total of 38 studies have been selected from 4 different academic databases. The selected studies have been reviewed based on five key aspects: dataset, preprocessing techniques, feature extraction techniques, classification methods and evaluation metrics. This review basically focuses on publications that have used discharge summary or other medical reports along with discharge summary as the dataset. The majority of studies (n=30) have used publicly available datasets MIMIC-II and MIMIC-III that are coded using ICD-9-CM codes. Various preprocessing techniques have been applied to remove unwanted or meaningless information from the discharge summaries which has helped to obtain improved classification results. In the majority of the studies, Word2Vec embedding and TF-IDF feature representation techniques have been determined to be beneficial. For classification, the  studies have used machine learning and deep learning approaches. The majority of studies have compared the performance of their deep learning model with the state-of-the-art method and provided improved classification results. In this review, we have compared the selected studies based three key aspects: dataset, train-test split size, and number of codes. In evaluation metrics, the majority of studies have used micro- and macro-averaging of precision, recall, and F1-score. Lastly, we have addressed various future research directions where a considerable efforts are required in order to develop an automated ICD coding system.

%\section*{Funding}

\section*{Conflict of interest}
The authors declare that they have not conflict of interest.

%\section*{Ethical approval}

%\section*{Informed consent}

\section*{Acknowledgement}
This work is supported by Western Sydney University Post Graduate Research Scholarship.

\printcredits

\appendix
%\section{My Appendix}
%Appendix sections are coded under \verb+\appendix+.

\bibliographystyle{cas-model2-names}

\bibliography{cas-refs}

%\begin{comment}
%Author biography without author photo.

%\bio{figs/pic1}
%Author biography with author photo.
%\end{comment}

\end{document}